\documentclass[twoside,11pt]{article}

\usepackage{jmlr2e}
\usepackage[utf8]{inputenc} 
\usepackage[T1]{fontenc}    
\usepackage{hyperref}       
\usepackage{url}            
\usepackage{booktabs}       
\usepackage{amsfonts}       
\usepackage{nicefrac}       
\usepackage{microtype}      
\usepackage{amsmath}
\usepackage{algorithm}
\usepackage{algorithmic}
\usepackage{graphicx}
\usepackage{multirow}
\usepackage{subcaption}
\usepackage{amssymb}
\newcommand*{\QEDB}{\hfill\ensuremath{\square}}

\jmlrheading{1}{2020}{submiited}{00/00}{00/00}{Huang2020}{Yimin Huang, Yujun Li, Hanrong Ye, Zhenguo Li and Zhihua Zhang}

\ShortHeadings{An Asymptotically Optimal MAB Algorithm and HPO}{Huang, Li, Li and Zhang}
\firstpageno{1}

\begin{document}

\title{An Asymptotically Optimal Multi-Armed Bandit Algorithm and Hyperparameter Optimization}

\author{\name Yimin Huang \email yimin.huang@huawei.com \\
       \addr Huawei Noah's Ark Lab\\
       Beijing 100871, China
       \AND
       \name Yujun Li \email liyujun145@gmail.com \\
       \addr Department of Computer Science\\
       Shanghai Jiao Tong University\\
       Shanghai 200240, China
	   \AND
	   \name Hanrong Ye \email ye.hanrong@huawei.com \\
	   \addr Huawei Noah's Ark Lab\\
	   Hong Kong, China
		\AND
	   \name Zhenguo Li \email Li.Zhenguo@huawei.com \\
	   \addr Huawei Noah's Ark Lab\\
	   Hong Kong, China
	   \AND
	   \name Zhihua Zhang \email zhzhang@math.pku.edu.cn \\
	   \addr School of Mathematical Sciences\\
	   Peking University\\
	   Beijing 100871, China}

\editor{ }

\maketitle

\begin{abstract}
	The evaluation of hyperparameters, neural architectures, or data augmentation policies becomes a critical model selection problem in advanced deep learning with a large hyperparameter search space. In this paper, we propose an efficient and robust bandit-based algorithm called Sub-Sampling (SS) in the scenario of hyperparameter search evaluation. It evaluates the potential of hyperparameters by the sub-samples of observations and is theoretically proved to be optimal under the criterion of cumulative regret. We further combine SS with Bayesian Optimization and develop a novel hyperparameter optimization algorithm called BOSS. Empirical studies validate our theoretical arguments of SS and demonstrate the superior performance of BOSS on a number of applications, including Neural Architecture Search (NAS), Data Augmentation (DA), Object Detection (OD), and Reinforcement Learning (RL). 
\end{abstract}




\section{Introduction}\label{sec:intro}

Hyperparameter optimization, a classical problem for model selection, receives increasing attention due to the rise of automated machine learning, where not only traditional hyperparameters such as learning rates \citep{zeiler2012adadelta}, but also neural architectures \citep{liu2018darts}, data augmentation policies \citep{cubuk2018autoaugment} and other plausible variables are to be set automatically \citep{elsken2019neural,zoller2019survey}. This can be perceived as a search problem in a large hyperparameter search space -- the goal is to find the hyperparameters that maximize the generalization capability of the model if trained with the searched hyperparameters. There are mainly three challenges: the initial design \citep{jones1998efficient,konen2011tuned,brockhoff2015impact,zhang2019deep}, the sampling method \citep{rasmussen2003gaussian,Hutter2012Sequential,Bergstra2011Algorithms,springenberg2016bayesian,srinivas2010gaussian,hennig2012entropy,hernandez2014predictive,wang2017max,ru2018fast}, and the evaluation method \citep{Thornton2013Auto,karnin2013almost,jamieson2016Non,li2016hyperband,falkner2018bohb}.

In the literature, initial designs including Latin hypercube design \citep{jones1998efficient} and orthogonal array \citep{zhang2019deep} are used to replace random initialization for improving experimental effect and reducing the time required for convergence. One straightforward idea of sampling methods is to use grid search 
\citep{wu2011experiments} or random search \citep{Bergstra2012Random}. However, the required number of function evaluations, which involve model training and thus computationally expensive, grows exponentially with the number of hyperparameters. 
Population-based methods \citep{Hansen2016cma,Jaderberg2017Population} such as genetic algorithms, evolutionary strategies, and particle swarm optimization use guided search and usually outperform random search. 
These algorithms are often time-consuming for the evolution of the population.
To develop efficient sampling strategies, model-based methods are paid more attention in which a surrogate model of the objective function is built, allowing to choose the next hyperparameters to evaluate in an informed manner. Among them, Bayesian optimization (BO) becomes popular with different probabilistic surrogate models, such as
Gaussian processes (GPs) \citep{Snoek2012Practical}, random forests \citep{Hutter2012Sequential}, or Tree-structure Parzen Estimator (TPE) \citep{Bergstra2011Algorithms}.
These model-based methods are shown to outperform random search in terms of sample efficiency \citep{Thornton2013Auto,snoek2015scalable}.
In this paper we focus on the  hyperparameter evaluation problem. In many machine learning problems such as Neural Architecture Search (NAS) and Data Augmentation (DA), neural architectures and augmentation policies can always be parameterized. In most scenarios, the performance of a hypothesis relies heavily on these hyperparameters. And each evaluation of hyperparameter configurations is expensive. Thus, an efficient evaluation method is a key point in hyperparameter optimization (HPO).
A general solution for fast, accurately and robustly evaluating these hyperparameters, neural architectures or augmentation policies from a large search space is desirable.

%

While the many previous studies focus on sampling strategies, multi-fidelity methods concentrate on accelerating function evaluation which could be extremely expensive for big data and complex models \citep{krizhevsky2012imagenet}.
These methods use low-fidelity results derived with small amount of resources to select configurations quickly. Details refer to \citet{feurer-automlbook18a}.
The successive halving (SH) algorithm \citep{jamieson2016Non} is proposed to identify the best configuration among $K$ configurations. It evaluates all hyperparameter configurations, throws the worst half, doubles the budgets and repeats until one configuration left. However, SH allocates exponentially more resources to more promising configurations. Besides, the number of configurations $K$ is required as an input, but there is no guidance for choosing a proper $K$. 
HyperBand (HB) \citep{li2016hyperband} addresses this trade-off by implementing the SH algorithm with multiple values of $K$, with each such run of SH as a ``bracket''.
\citet{falkner2018bohb} further combined Bayesian optimization (BO) and HyperBand (HB), termed BOHB, which extracted the information in the current bracket to fit a Bayesian surrogate model for sampling configurations in the next bracket. So far, there is a complete solution to the problem of slow evaluation of hyperparameter optimization. However, it is worth noting that the HB algorithm was proposed for identifying the best configuration among the alternative set, not for collecting data to estimate the sampling model in BO. Hence, combining BO and HB directly is not appropriate, which will be addressed in our work. These highly relevant works will be explained in more detail in Section \ref{sec:related}.  
Recently, there many interesting work about HPO. \citet{paul2019fast} proposed an algorithm for tuning hyperparameters of the policy gradient methods in RL.
\citet{keshtkaran2019enabling} tuned hyperparameters of a sequential autoencoder for spiking neural data. \citet{law2019hyperparameter} learned a joint model on hyperparameters and data representations from many previous tasks.
\citet{li2020rethinking} considered hyperparameter selection when fine-tuning from one domain to another. \citet{klein2019meta} presented a meta-surrogate model for task generation for HPO which facilitates developing and benchmarking of HPO algorithms. 

\begin{figure}[ht]
	\begin{center}
		\includegraphics[width=0.48\textwidth]{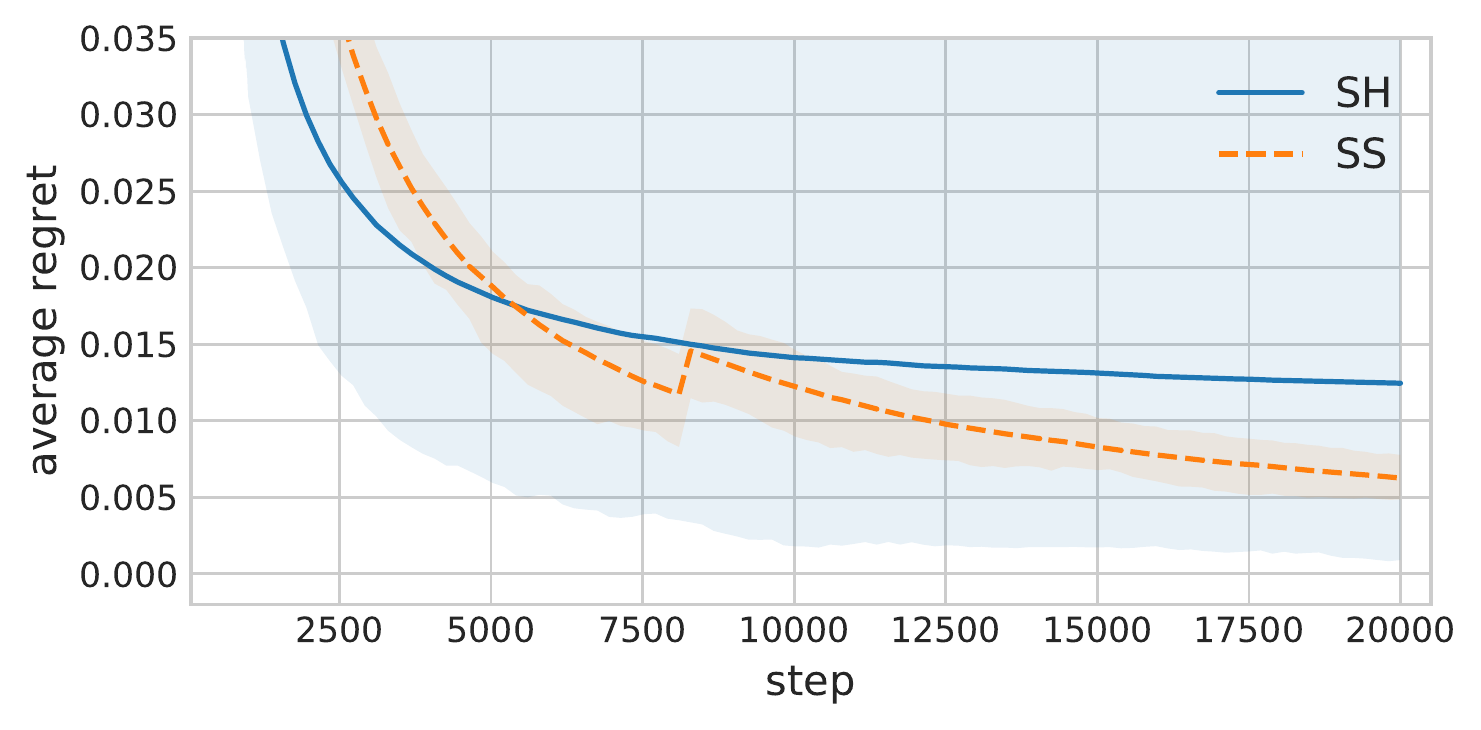}
		\caption{Using SS and SH as policies for Multi-Armed Bandit (MAB) problems (Section \ref{sec:sim}). The shaded part indicates the min/max ranges.
		}
		\label{fig:show}
	\end{center}
\end{figure}

In this paper we focus on the need for collecting high-quality data with multi-fidelity methods. We propose a new efficient and general-purpose algorithm for fast evaluation called Sub-Sampling (SS). Compared with the HB algorithm, SS evaluates the potential of the configurations based on the sub-samples of observations. It does not focus on finding the best configuration only as in SH, but instead guarantees the perfect overall performance in the sense that the cumulative regret of SS is asymptotically optimal. Section \ref{sec:theory} provides theoretical analysis and Figure \ref{fig:show} compares SS and SH on one example. SS has a jump in the figure since its criterion is based on the potential rather than the current performance of the configurations.

Further, we combine BO and SS to deal with many popular machine learning tasks in Section \ref{sec:app}. The estimation of the surrogate model in the BO framework is more accurate based on these promising data obtained by SS. In the next bracket, we can sample better configurations from the model with higher probability. As a result, the trajectory of the search procedure is more reliable. Extensive experiments on various problems show the superior performance of BOSS, including Neural Architecture Search (NAS), Data Augmentation (DA), bounding box scaling in Object Detection (OD), and tuning hyperparameters of PPO \citep{schulman2017proximal} in Reinforcement Learning (RL).
\section{Related Work}\label{sec:related}
In this section, we will provide a slice of details regarding the multi-fidelity methods. The shortcomings of historical methods will be revealed. 

\subsection{Successive Halving}
The basic objective of multi-fidelity methods is to identify the best configuration out of a given finite set of configurations based on low-fidelity approximations of their performance. In order to achieve this objective, it is possible to drop hyperparameter configurations if they perform badly with small computing resources. Based on this idea, \citet{jamieson2016Non} proposed the Successive Halving algorithm originally introduced by \citet{karnin2013almost} for HPO. Query all $K$ configurations with a given initial budget for each one; then, remove the half that performed worst, double the
budget and successively repeat until only a single configuration is left, which is described in Algorithm \ref{alg:sh}. The main theorem in \citet{jamieson2016Non} indicates that SH can return the true best configuration when the total used budget $B$ is larger than a certain value. SH is an extremely simple, yet powerful, and therefore popular policy for the selection of multi-fidelity methods. \citet{li2018massively} presented Asynchronous Successive Halving Algorithm (ASHA), a practically parallelized version of SH for dealing that configurations are typically orders of magnitude more than available parallel workers.

\begin{algorithm}
	\caption{Successive Halving}
	\label{alg:sh}
	\textbf{Input:} The set of $K$ configurations $\mathcal{C}=\{c_1,\ldots,c_K\}$; minimum budget $b$; ratio $\eta$.\\
	\textbf{Output:}  Evaluation results of configurations.
	
	\begin{algorithmic}[1]
		\STATE $s=\lfloor\log_{\eta}(K)\rfloor$, total budget $B=Kbs$
		\FOR{$r=0$,..., $s$}
		\STATE $K_r=\lfloor K\eta^{-r}\rfloor$
		\STATE $b_r=b\eta^{r}$
		\STATE Evaluate current $K_r$ configurations with budgets $b_r$
		\STATE Keep top $K_r/\eta$ configurations
		\ENDFOR
	\end{algorithmic}
\end{algorithm}

As discussed in Section \ref{sec:intro}, SH suffers from the trade-off between the budget and the number of configurations. Given a total budget of $B=Kbs$, users need to decide the number of configurations $K$. Trying larger $K$ and assigning a small budget to each can result in prematurely terminating good configurations while trying only a few and assigning them a larger budget can result in wasting resources on evaluating a poor configuration.

\subsection{HyperBand}
Hyperband \citep{li2016hyperband} is presented to combat the trade-off problem in SH. It uses different values of $K$ and calls the SH algorithm as a subroutine. There are two components to HB shown in Algorithm \ref{alg:hb}; (1) the inner loop invokes SH for fixed values of $K$ and $b$; (2) the outer loop iterates over different values of $K$ and $b$.

\begin{algorithm}
	\caption{HyperBand}
	\label{alg:hb}
	\textbf{Input:} Maximum budget $R$; ratio $\eta$\\
	\textbf{Output:} The configuration with the best performance
	
	\begin{algorithmic}[1]
		\STATE $s_{max}=\lfloor\log_{\eta}(R)\rfloor$, $B=(s_{max}+1)R$
		\FOR{$s=s_{max},s_{max}-1,\ldots,0$}
		\STATE $K=\lceil \frac{B\eta^s}{R(s+1)}\rceil$, $b=R\eta^{-s}$
		\STATE Sample $K$ configurations randomly
		\STATE Call SH with $(K,b)$
		\ENDFOR
	\end{algorithmic}
\end{algorithm}

HyperBand begins with the most aggressive bracket $s = s_{max}$, which sets $K$ to maximize exploration, subject to the constraint that at least one configuration is allocated $R$ resources. Each subsequent bracket reduces $K$ by a factor of approximately $\eta$ until the final bracket, $s=0$, in which every configuration is allocated $R$ resources (this bracket simply performs the classical random search). 

In practice, HB works very well and typically outperforms random search and Bayesian optimization methods for small total budgets. However, the brackets in HB are independent of each other. And, the configurations in the next bracket are still sampled randomly without any guidance which means the information of previous brackets wastes.

\subsection{Bayesian Optimization and HyperBand}

To overcome the limitation that HB does not adapt the configuration sampling methods to the evaluations, the recent approach BOHB \citep{falkner2018bohb} combines Bayesian optimization and HyperBand. Its idea is to build the relation between brackets by the Bayesian model and to replace HB's random search by BO. Particularly, as shown in Algorithm \ref{alg:bohb}, BOHB relies on HB to determine how many resources used to evaluate configurations, but it replaces the random selection of configurations in each bracket by a model-based search. Once the number of configurations in a bracket is determined, the standard SH algorithm runs with these configurations. Then, we collect these configurations and their performance to fit a Bayesian surrogate model. For the next bracket, the configurations will be sampled from the model. Compared to the original HB, BOHB guides the search. There are two other works that also attempted to combine Bayesian optimization with HyperBand. \citet{bertrand2017hyperparameter} used a Gaussian process with a squared exponential kernel as the Bayesian surrogate model instead of the TPE method of BOHB and dealt with model selection tasks. \citet{wang2018combination} sampled trial points one by one using BO building the relation of configurations in the bracket, not between brackets.

\begin{algorithm}
	\caption{BOHB}
	\label{alg:bohb}
	\textbf{Input:} $R$, the maximum amount of resources that can
	be allocated to a single configuration; ratio $\eta$\\
	\textbf{Output:} The configuration with the best performance 
	
	\begin{algorithmic}[1]
		\STATE $s_{max}=\lfloor\log_{\eta}(R)\rfloor$, $B=(s_{max}+1)R$
		\FOR{$s=s_{max},s_{max}-1,\ldots,0$}
		\STATE $K=\lceil \frac{B\eta^s}{R(s+1)}\rceil$, $b=R\eta^{-s}$
		\STATE Sample $K$ configurations from the Bayesian model
		\STATE Call SH with $(\mathcal{C},b)$
		\STATE Use the data from SH to refit the model
		\ENDFOR
		
	\end{algorithmic}
\end{algorithm}

Unfortunately, HB is designed to identify the best configuration. There is no guarantee for the evaluation of other configurations while BO needs that all configurations not only the best one used to estimate the model have high-level performance. In other words, it will lead to a wrong estimation of the surrogate model. Therefore, it is necessary to propose another criterion for fast evaluation.

\subsection{Bandit-based Methods concerning Cumulative Regret}
In the literature, there are many bandit policies proposed to get optimal cumulative regret, which means they pursue an optimal sequence of configurations instead of the final return configuration only. \citet{lai1985asymptotically} gave an asymptotic lower bound for the regret in the multi-armed bandit problem and proposed an index strategy that achieved this bound. \citet{lai1987adaptive} showed that when probability distributions belong to a specified exponential family, a policy that pull the arm of the largest upper confidence bound (UCB) is optimal. The UCB policy is constructed from Kullback-Leibler (KL) information between estimated observation distributions of the arms.
\citet{agrawal1995sample} modified the UCB policy without knowing the total sample size. To better describe applications, the first work to venture outside the realm of parametric modeling assumptions
appeared in \citet{Yakowitz1991Nonparametric}. As opposed to the traditional multi-armed bandit problem, they proposed non-parametric policies not based on KL-information but under some moment conditions. \citet{auer2002finite-time} provided the policy named UCB$1$ which can achieve logarithmic regret if observation distributions are supported on $[0,1]$. \citet{Chan2019The} proposed an efficient non-parametric solution and proved optimal efficiency of the policy. However, their observation distribution must belong to an one-parameter exponential family which would be extended in our work. Moreover, they only handled the standard multi-armed bandit problems while HPO is the ultimate goal in our work.

\section{Multi-Armed Bandit Problem for HPO}\label{sec:mab}

In this section the notation and the standard multi-armed bandit problem with $K$ configurations (arms) will be introduced for discussing another criterion, cumulative regret. 

Recall the traditional setup for the classic multi-armed bandit problem. Let $\mathcal{I}=\{1,2,\ldots,K\}$ be a given set of $K\geq2$ configurations. Consider a sequential procedure based on past observations $Y_{t}^{(k)}$, where $t$ represents the observation time from the $k$-th configuration. Let $N_{k}$ be the number of observations from the $k$-th configuration with different budgets, and $N=\sum_{k=1}^{K}N_{k}$ is the number of total observations. For each configuration, the observations $\{Y_{t}^{(k)}\}_{t\geq1}$ are assumed to be independent and identically distributed with expectation given by $\mathbb{E}(Y_{t}^{(k)})=\mu_{k}$ and $\mu_{*}=\min_{1\leq k\leq K}\mu_{k}$. In HPO problems, the randomness of the observations comes from the randomness of initialization, because the performance of an under-trained neural network with small budgets strongly depends on its initialization.
For simplicity, assume without loss of generality that the best configuration is unique which is also assumed in \citet{perchet2013the} and \citet{Chan2019The}.

A policy $\pi=\{\pi_{t}\}$ is a sequence of random variables $\pi_{t}\in\{1,2,\ldots,K\}$ denoting that at each time $t=1,2,\ldots,N$, the configuration $\pi_{t}$ is selected to evaluate. Note that $\pi_{t}$ depends only on previous $t{-}1$ observations. The objective of a good policy $\pi$ is to minimize the cumulative regret 
$$R_{N}(\pi)\triangleq\sum_{k=1}^{K}(\mu_{k}-\mu_{*})\mathbb{E}N_{k}=\sum_{t=1}^{N}(\mu_{\pi_{t}}-\mu_{*}).$$
Note that for a data-driven policy $\hat{\pi}$, the regret monotonically increases with respect to $N$. Hence, minimizing the growth rate of $R_{N}$ is an important criterion which is considered in the later section. The successive elimination method  \citep{perchet2013the} is like SH to eliminate bad configurations successively and is given an upper bound of the cumulative regret for any $N$. However, the asymptotic property that we need in big data is not optimal.

An arm allocation policy $\pi$ is said to be uniformly good if 
$$R_N(\pi)=o(N^\varepsilon) {\text{ for all }}\varepsilon>0.$$
Moreover, if $\pi$ is uniformly good and some additional regularity conditions are satisfied, \citet{lai1985asymptotically} provided a lower bound of the growth rate:
\begin{equation}\label{equ:bound}
\liminf_{N\rightarrow\infty}\frac{R_N(\pi)}{\log N}\geq\sum_{k:\mu_*<\mu_k}\frac{\mu_k-\mu_*}{KL(f_k,f_*)}.
\end{equation}
$KL(f,g)$ is the Kullback-Leibler divergence between density functions $f$ and $g$, that is,
$$KL(f,g)=\mathbb E_f\Big[\log\frac{f(Y)}{g(Y)}\Big],$$
where $\mathbb E_f$ denotes expectation with respect to $Y\sim f$.
\citet{Chan2019The} proposed an arm allocation policy in a strong assumption on the distribution of reward $Y$ that made regrets achieve the lower bound in Equation (\ref{equ:bound}). We say that it has the optimal rate. And, if the growth of the cumulative regret has the order of $\log N$, it is called a nearly optimal policy.

\section{Proposed Sub-Sampling for Efficient Hyperparameter Evaluation}\label{sec:ss}

In this section, we propose a novel efficient nonparametric solution called Sub-Sampling (SS) for evaluating a pool of configurations $\mathcal{C}=\{c_1,\ldots,c_K\}$ that minimizes the total regrets.
The main idea is to assign more budgets to the configuration of more potential. Let $Y_t^{(k)}$ be the validation loss of $c_k$ at the $t$-th evaluation, $n_k$ be the number of evaluations of $c_k$ so far, and $n=\sum_{i=1}^Kn_k$ be the total number of evaluations for all the $K$ configurations.
To compare configurations for selecting next ones to evaluate with designed budgets, SS uses all available data $\{Y_t^{(k)}\}_{t=1}^{n_{k}}$ generated so far from different rounds with different initializations\footnote{$Y_t^{(k)}$ is the accuracy of the corresponding network trained with the certain budgets used in this round and the $t$-th initialization.} to measure the potential of configurations while SH just uses the data $Y_{n_k}^{(k)}$ from the current round. Note that in the early stage of network training, the performance relies strongly on the initialization which has high randomness. SS uses data which has $n_k$ samples of $c_k$ for each $k$ to reduce the impact of randomness. This reduces probability of misjudging in SS.

Specifically, $c_k$ is said to has more potential than $c_{k'}$, denoted by $c_k\preccurlyeq c_{k'}$, if $n_{k}<n_{k'}$ and (a) $n_{k}<q_n$ or (b) $q_n\leq n_{k}$ and $\bar{Y}_{1:n_{k}}^{(k)}\leq\bar{Y}_{j:(j+n_{k}-1)}^{(k')}$, for some $1\leq j\leq n_{k'}-n_{k}+1$,
where $\bar{Y}_{l:u}^{(k)}=\sum_{v=l}^{u}Y_{v}^{(k)}/(u-l+1)$.
In this definition, $n_k$ sub-samples from $n_{k'}$ samples of $c_{k'}$ are used to compare $c_k$ and $c_{k'}$, thus called sub-sampling. 
The parameter $q_n$ is given to balance exploration and exploitation. 
It is a nonnegative increasing threshold for SS such that $q_n=o(\log n)$ and $q_n/\log\log n\rightarrow\infty$ as $n\rightarrow\infty$.


In case (a), it reveals that the performance of $c_k$ is observed too little, so we cannot judge its potential. Thus, $c_k$ needs to be explored. In case (b), it indicates that $c_k$ has the potential to exceed $c_{k'}$, although the current performance $Y^{(k)}_{n_k}$ may be worse than $Y^{(k')}_{n_{k'}}$. Hence, $c_k$ needs to be exploited with more budgets.
\begin{algorithm}[ht]
	\caption{\label{alg:SS} Sub-Sampling}
	\textbf{Input:} The set of $K$ configurations $\mathcal{C}=\{c_1,\ldots,c_K\}$; maximum budget $R$; minimum budget $b$;  ratio $\eta$ (default $\eta=3$).\\
	\textbf{Output:} $\{c_{\hat{\pi}_{1}},\ldots,c_{\hat{\pi}_{N}}\}$ with corresponding evaluations. 
	
	\begin{algorithmic}[1]
		\STATE $r=1$, evaluate all configurations with budget $b$.
		\FOR{$r=2,3,\ldots,[\log_{\eta}(R/b)]$} 
		\STATE Select the leader $c_\zeta$, which has the most observations. 
		\STATE $\mathcal{I'}=\{k:c_k\in\mathcal{C}\backslash c_\zeta,\ c_k\preccurlyeq c_{\zeta}\}$.
		\IF {$\mathcal{I'}==\varnothing$}
		\STATE Evaluate $c_\zeta$ with budgets $\eta^rb$.
		\ELSE
		\STATE Evaluate $c_k$ with budgets $\eta^rb$ for each {$k\in\mathcal{I'}$}. 
		\ENDIF
		\ENDFOR
		
	\end{algorithmic}
\end{algorithm}

The proposed SS method is described in Algorithm \ref{alg:SS}.
The sequence of the configurations $\hat{\pi}_{1},\ldots,\hat{\pi}_{N}\in\mathcal{I}=\{1,2,\ldots,K\}$ is a sampling strategy for HPO, where $N=\sum_{k=1}^{K}N_{k}$ is the number of total observations. This strategy $\hat\pi=\{\hat\pi_{t}\}$ is a sequence of random variables $\hat\pi_{t}\in\{1,2,\ldots,K\}$ denoting that at each time $t=1,2,\ldots,N$, the $\hat\pi_{t}$-th configuration is selected to evaluate. Let $r$ denote the round number. In the first round, all configurations are evaluated with minimum budget $b$ since there is no information about them. In round $r\geq2$, we select the leader of configurations which is the one evaluated the most times, and the budgets increase as $\eta^rb$. If two configurations have the same number of observations $n_k$, we choose the one with lower $\bar Y_{1:n_k}^{(k)}$ as the leader. In each round $r\geq2$, non-leaders will be evaluated, if they have more potential than the leader, otherwise, the leader will be evaluated. 

\textbf{SS vs SH}\\
In HPO problems, the randomness of the early observations comes from the randomness of initialization, since the performance of an under-trained neural network with small budgets strongly depends on its initialization. Thus, the comparisons at the first stage of SH are unreliable and the abandoned configurations have no chance to be evaluated again. In contrast, SS uses data of all stages to weaken the impact of initial value and always gives a chance for each configuration to make the comparison of potential. Consequently, SS can get more reliable results. In Section \ref{sec:sim}, Table \ref{tab:SS_acc} shows that SS has higher accuracy of finding the optimal configuration than SH especially in the case of big randomness.

Both SS an SH are different from the UCB-based procedures \citep{burtini2015survey} which must know the underlying probability distributions to measure the potential of a configuration by an upper confidence bound of the observation value. However, only SS achieves asymptotically optimal efficiency which will be discussed in Section \ref{sec:theory}.

It should be noted that sub-sampling is also used in \citet{Chan2019The} for multi-armed bandit (MAB). One key difference is that different comparing criteria for different distributions are needed in \citet{Chan2019The} while we can use the same criterion for these distributions and achieve the optimality (Section \ref{sec:theory}).

\section{Theoretical Results}\label{sec:theory}
In this section, we prove that the proposed SS method is asymptotically optimal using the tools from multi-armed bandit (MAB) \citep{agarwal2012oracle,sparks2015tupaq,jamieson2016Non}. Each arm corresponds to a fixed hyperparameter setting, the arm collection $\mathcal{I}$ corresponds to the set of configurations $\mathcal{C}$, pulling an arm corresponds to a fixed number of training iterations, budget corresponds to the number of samples in one pull and the loss corresponds to the validation loss.

Given an arm allocation policy $\pi$, consider the cumulative regret $R_N(\pi)=\sum_{k=1}^{K}(\mu_{k}-\mu_{*})\mathbb{E}N_{k}$, where $\mu_{k}=\mathbb{E}(Y_{t}^{(k)})$ and $\mu_{*}=\min_{1\leq k\leq K}\mu_{k}$. 
\citet{lai1985asymptotically} provided an asymptotic lower bound of $R_N(\pi)/\log N$ with $\sum_{k:\mu_*<\mu_k}{(\mu_k-\mu_*)}/{KL(f_k,f_*)}$,
where $f_k(y)$ is a density function of $Y_t^{(k)}$, $f_*=f_{\arg \min_k \mu_k}$, and the function $KL(\cdot,\cdot)$ is the Kullback-Leibler divergence.
The cumulative regret is called near optimality if $R_N(\pi)=O(\log N)$, or optimality if it achieves this optimal growth rate specified by the lower bound.

In the literature, most researchers only considered a one-parameter exponential family \citep{perchet2013the,Chan2019The}.
For wider applications, we study a general exponential family defined by
\begin{equation}\label{exp}
f(y,\theta,\phi)=e^\frac{\theta y-g(\theta)}{\phi}h(y,\phi),\ \phi>0,
\end{equation}
where functions $g$ and $h$ are decided by a specific probability distribution. When $h(y,\phi)=1/\phi$, it is a standard exponential distribution. The normal distributions $N(\mu,\sigma^2)$ are included with $\theta=\mu,\phi=\sigma^2,g(\theta)=\theta^2/2,h(y,\phi)=y^2/(2\phi)+0.5\log(2\pi\sigma^2)$. Let $f_{k}(y)=f(y,\theta_k,\phi_k)$ for $k=1,\ldots,K$.
Let $f_{*}=f(\cdot,\theta_*,\phi_*)$, where $\mu_*=\min_{1\leq k\leq K}\mu_k$ and $\theta_*$, $\phi_*$ are the corresponding parameters. Note that $\mu=\mathbb E Y=g'(\theta)$ and $Var(Y)=\phi g''(\theta)>0$. Let $b(\mu)$ be the inverse function of $g'(\theta)$. The following theorem gives the property of the proposed sub-sampling method (Algorithm \ref{alg:SS}) for the exponential family. The near optimality is obtained by bounding the tails of distributions. The proof is given in Appendix.

\begin{theorem}\label{thm:two1}
	For the exponential family (\ref{exp}), the SS policy $\hat\pi$ given in Algorithm \ref{alg:SS} satisfies
	\begin{align*}
	\limsup_{N\rightarrow\infty}\frac{R(\hat{\pi})}{\log N} \leq\sum_{k:\mu_{*}<\mu_{k}}\frac{(\mu_{k}-\mu_{*})\phi_{*}}{(b(\mu_{k})-b(\mu_{*}))\mu_{k}-(g(b(\mu_{k}))-g(b(\mu_{*})))},
	\end{align*}
	and is thus nearly optimal.
\end{theorem}

Since the large deviation rate function is not $KL$ divergence under the exponential family, this upper bound is not optimal. But, it has the nearly optimal order of $\log N$. Consider the classical case of the one-parameter exponential family, i.e., $\phi_k=1$ for any $k$, the large deviation rate function $I_*(\mu_k)$ turns out to be $KL$ divergence by direct calculation. It means that the sub-sampling method is optimal under the one-parameter exponential family \citep{Chan2019The}. Corollary \ref{thm:one} reveals that the proposed policy is optimal for the one-parameter exponential family. The upper bound of the regret given in \citet{perchet2013the} does not have this optimality. 

\begin{corollary}\label{thm:one}
	For the one-parameter exponential family, the SS policy $\hat\pi$ given in Algorithm \ref{alg:SS} satisfies
	$$\limsup_{N\rightarrow\infty} \frac{R(\hat\pi)}{\log N}\leq\sum_{k:\mu_*<\mu_{k}}\frac{\mu_{k}-\mu_*}{KL( f_{k},f_{*})},$$
	and is thus optimal.
\end{corollary}

From these two theoretical results, it reveals that when the distributions of different arms are more different, the convergence rate is faster since we can identify the best arm quickly. And, smaller $\mu_k-\mu*$ will naturally lead to smaller regret.

\section{Bayesian Optimization via Sub-Sampling}\label{sec:boss}
In this section, we propose a novel algorithm called BOSS, which combines Bayesian Optimization (BO) and Sub-Sampling (SS), to search out the optimal configurations efficiently and reliably.  

Bayesian optimization is a sequential design strategy for optimizing black-box functions. In hyperparameter optimization problems, the validation performance of a machine learning algorithm can be regarded as a function $f:\mathcal X\rightarrow\mathbb{R}$ of hyperparameters
$x\in\mathcal X$ and the goal is to find the optimal $x_{*}\in \arg\min_{x\in\mathcal X}f(x)$. In most cases, $f(x)$ does not admit an analytic form, which is approximated by a surrogate model in BO, such as Gaussian processes, random forests, or TPE, based on the data collected on the fly $D_n=\{(x_1,y_1),\ldots,(x_n,y_n)\}$ where $y_i=f(x_i)+\varepsilon(b)$ and the error term $\varepsilon(b)$ is dependent on the budget $b$ and satisfies that $\lim_{b\rightarrow\infty}\varepsilon(b)=0$. The standard algorithmic procedure of BO is stated as follows.
\begin{enumerate}
	\setlength{\itemsep}{0pt}
	\setlength{\parsep}{0pt}
	\setlength{\parskip}{0pt}
	\item Assume an initial surrogate model.
	\item Compute an acquisition function $a:\mathcal X\rightarrow\mathbb R$ based on the current model.
	\item Sample a batch of hyperparameter configurations based on the acquisition function.
	\item Evaluate the configurations and refit the model.
	\item Repeat steps 2-4 until the stop condition is reached.
\end{enumerate}
The BO methods differ in different surrogate models for the conditional probability $p(y|x,D_n)$. In this work, we adopt TPE \citep{Bergstra2011Algorithms} as the surrogate model, which utilizes a kernel density estimator to model the data densities to deal with both discrete and continuous hyperparameters simultaneously. The specific procedure of TPE with the expected improvement (EI) as the acquisition function is given as follows.

This strategy models $p(x|y,D_n)$ instead of $p(y|x,D_n)$ and uses two densities below to estimate:
\begin{equation} \label{eq:tpe}
p(x|y,D_n)=\left\{ 
\begin{aligned}
\ell(x|D_n) &\quad \text{if}\quad y<\alpha \\
g(x|D_n) &\quad \text{if}\quad y\geq\alpha,
\end{aligned}
\right.
\end{equation}
where $\alpha$ is a given demarcation point and its value will be discussed later. $\ell(x)$ is the density satisfying that the observation $y$ was less than $\alpha$ and is assumed to be irrelevant with the specific value of $y$. It is estimated by a kernel density estimator with the observations $\{x_i\}$ such that corresponding observation $y_i$ was less than $\alpha$. And, $g(x)$ is the similar density formed by using the remaining observations.

According to the model, we set an acquisition function to sample next point $x$ that is most likely to be optimal. A common acquisition function is the expected improvement (EI):
\begin{equation}\label{eq:ei}
a_\alpha(x)=\mathbb E_{p(y|x,D_n)}(\alpha-y)\mathbb I(y\leq\alpha),
\end{equation}
where $\mathbb I$ is the indicator function.

\citet{Bergstra2011Algorithms} showed that maximizing the EI function in Equation (\ref{eq:ei}) is equivalent to maximizing the ratio $\ell(x|D_n)/g(x|D_n)$ of densities in Equation (\ref{eq:tpe}).

Note that the observations usually represent the value of a loss function in model training. We pay more attention in the area of lower losses. Hence, the estimation of $\ell(x)$ is much more vital. So, we want smaller $\alpha$ to be better. Unfortunately, if $\alpha$ is too small, there is no enough data for estimation. In practice, the TPE algorithm chooses $\alpha$ to be some quantile $\gamma$ of the observed values.

In the literature of BO, surrogate models (step $1$) \citep{springenberg2016bayesian}, acquisition functions (step $2$) \citep{wang2017max} and batch sampling methods (step $3$) \citep{gonzalez2016batch} are well studied, but the problem of hyperparameter configuration's evaluation (step $4$) remains largely open. While one could evaluate a hyperparameter configuration by training the corresponding model until convergence, this is quite expensive for complex models and big data especially when there are many configurations to evaluate. In BOHB, this issue is addressed with SH. However, SH only guarantees the performance of the best configuration, implying that the total data used to estimate the surrogate model can be extremely poor. In multi-armed bandit, total regrets are more desired than the regret of the best one, which is the key motivation in our work. 

The proposed algorithm BOSS is described in Algorithm \ref{alg:BOSS}. The key in our design is a sub-sampling method for hyperparameter configurations evaluation, which we theoretically prove to be asymptotically optimal. 

\begin{algorithm}
	\caption{BOSS}
	\label{alg:BOSS}	
	\textbf{Input:} Maximum budget $R$; ratio $\eta$.\\
	\textbf{Output:} The configuration with the best performance. 
	
	\begin{algorithmic}[1]
		\STATE Initialize the surrogate model and the acquisition function with a uniform distribution.
		\STATE $s_{\max}=\lfloor\log_{\eta}(R)\rfloor$, $B=(s_{\max}+1)R$.
		\FOR{$s=s_{\max},s_{\max}-1,\ldots,0$}
		\STATE $K=\lceil \frac{B\eta^s}{R(s+1)}\rceil$, $b=R\eta^{-s}$.
		\STATE Sample $K$ configurations $\mathcal{C}$ from the acquisition function.
		\STATE Call SS with $(\mathcal{C},b,\eta)$.
		\STATE Use the output data from SS to refit the model and update the acquisition function.
		\ENDFOR
	\end{algorithmic}
\end{algorithm}

During initialization, we calculate the maximum stages through the maximum budget $R$ allowed for a single configuration and the ratio $\eta$. Then, the number of configurations $K$ and the minimum budget $b$ in each bracket are obtained. For initialization, we sample the configurations $\mathcal{C}$ from a uniform distribution and call the SS algorithm to evaluate $\mathcal{C}$. The collected data helps to refit the model and update the acquisition function for sampling in the next bracket. This BO framework gives the final best configuration.

\section{Asynchronous Parallelization}\label{sec:par}
In the BO framework, parallelization cannot happen between different brackets since the model needs to be updated. Hence, parallel acceleration occurs within each bracket. Both BOHB and BOSS can be only accelerated in SH or SS. The difference is that in SH, we know the number of configurations which need to be evaluated in the next round, so we can choose $K/\eta^r$ good configurations in advance. It is not necessary to wait until all configurations are evaluated. This asynchronous parallelization is a simple version of ASHA \citep{li2018massively} and implemented in BOHB. For SS, we do not know the number of configurations which need to be evaluated in the next round. This makes asynchronous parallelization like ASHA fails.

\begin{algorithm}
	\caption{\label{alg:pSS} Modified Sub-Sampling}
	\textbf{Input:} The set of $K$ configurations $\mathcal{C}=\{c_1,\ldots,c_K\}$; minimum budget $b$; conserved factor $\beta$; ratio $\eta$.\\
	\textbf{Output:} $\{c_{\widetilde{\pi}_{1}},\ldots,c_{\widetilde{\pi}_{N}}\}$ with corresponding evaluations. 
	
	\begin{algorithmic}[1]
		\STATE $s=\lfloor\log_{\eta}(K)\rfloor$, $V_k^{(-1)}=0$ for any $k\in\mathcal I.$
		\FOR{$r=0,\ldots,s$} 
		\STATE $K_r=\lfloor K\eta^{-r}\rfloor$, $b_r=b\eta^{r}$
		\STATE Evaluate top $K_r$ configurations with budgets $b_r$ sorted by $V_k^{(r-1)}$ in ascending order.
		\STATE Select the leader $c_\zeta$.
		\STATE Compute the sorting criterion $V_k^{(r)}(\zeta)$ for $k\in\mathcal I$.
		\ENDFOR
		
	\end{algorithmic}
\end{algorithm}

For the same reason, the total budget $B$ in BOHB is known while it is not clear in BOSS. It makes us only compare the two methods with the same maximum budget $R$, not with the same total budget. In order to give a fair comparison and for asynchronous parallelization, we propose a modified version.

We modify Algorithm \ref{alg:SS} to a new version by defining a criterion $V_k^{(r)}(\zeta)$ to sort these configurations in round $r$, where $V_k^{(r)}(\zeta)=\bar{Y}_{1:n_{k}}^{(k)}-\max_{1\leq j\leq n_{\zeta}-n_{k}+1}\bar{Y}_{j:(j+n_{k}-1)}^{(\zeta)}-\beta\cdot\max(0,q_n-n_k)$. 
If the conserved factor $\beta$ is larger, case (a) in Section \ref{sec:ss} will have a higher priority than case (b). 
The modified sub-sampling (MSS) algorithm is described in Algorithm \ref{alg:pSS}.

The procedure of MSS is similar to SH except that the sorting criterion is changed from $Y_{n_k}^{(k)}$ to $V_k^{(r)}$. Therefore, it is easy to utilize parallel resources like ASHA.
Finally, BOSS can be accelerated in parallel by replacing SS with MSS in step $6$ in Algorithm \ref{alg:BOSS}. 

\subsection{Aggressive Parallelization}

\begin{algorithm}[h]
	\caption{parallel BOSS}
	\label{alg:parallelBOSS}	
	\textbf{Input:} Maximum budget $R$; minimum budget $r$; ratio $\eta$; maximum duration $T$.\\
	\textbf{Output:} The configuration with the best performance. 
	
	\begin{algorithmic}[1]
		\STATE Initialize BOSS with $R$, $r$ and $\eta$.
		\WHILE{there is idle GPU}
		\IF{duration $t\geq T$}
		\STATE Break
		\ENDIF
		\STATE Find the next configuration $c_k$ and budget $b$ to evaluate based on Algorithm~\ref{alg:parallel_runtime}.
		\STATE Evaluate configuration $c_k$ with budget $b$ and update the Gaussian Process models.
		\ENDWHILE	
	\end{algorithmic}
\end{algorithm}

\begin{algorithm}[h]
	\caption{parallel BOSS runtime}
	\label{alg:parallel_runtime}
	\textbf{Input:} Current bracket $s$, current SS iteration $r$, maximum budget $R$.\\
	\textbf{Output:} The next configuration $c_k$ and budget $b$ to evaluate.
	
	\begin{algorithmic}[1]
		\IF{all configurations of current SS iteration $r$ scheduled}
		\IF{all configurations of current bracket $s$ scheduled}
		\STATE Sample a new bracket $s_{next}$, return one configuration $c_k$ of the first SS iteration, and $b=R\eta^{-s_{next}}$.
		\ELSE
		\IF{there is performace record of this bracket $s$}
		\STATE  Select the leader $c_\zeta$, which has the most observations.
		\STATE Return the top one configuration $c_k$ of the next SS iteration $r+1$ based on the sorting of $V_k^{(r)}(\zeta)$, and budget $b=R\eta^{-s+r+1}$.
		\ELSE
		\STATE Randomly pick one configuration $c_k$ from this bracket, with budget $b=R\eta^{-s+r+1}$.
		\ENDIF
		\ENDIF
		\ELSE
		\IF{there is performace record of this bracket}
		\STATE  Select the leader $c_\zeta$, which has the most observations. 
		\STATE Return the top one configuration $c_k$ of the current SS iteration $r$ based on the sorting of $V_k^{(r)}(\zeta)$, and budget $b=R\eta^{-s+r}$.
		\ELSE
		\STATE Randomly pick one configuration $c_k$ from this bracket, with budget $b=R\eta^{-s+r}$.
		\ENDIF
		\ENDIF
	\end{algorithmic}
\end{algorithm}

For the large-scale hyperparameter optimization problem for deep learning with GPU devices, the GPU computation resource is highly valuable. However, the difference of configuration numbers in the adjoining iterations and brackets results in idle GPUs from time to time, which reduces the actual searching time significantly and sometimes leads to inferior performance. 

To address this issue, we put forward an aggressive asynchronous parallel version of BOSS (``parallel BOSS'') as shown in Algorithm~\ref{alg:parallelBOSS} and Algorithm~\ref{alg:parallel_runtime}. The design philosophy of this parallel version is to minimize the GPU idle time. 
In this algorithm, when there is availabe GPU and returned evaluation results in the current bracket, it returns the top one configuration $c_k$ of the current SS iteration based on the sorting of $V_k^{(r)}(\zeta)$.
When there is no returned evaluation results in this bracket, the algorithm will randomly pick one configuration to evaluate.
In this way, the algorithm makes the best use of GPU resources.
On the other hand, the exploration of this algorithm is strengthened while the exploitation is weaken.

Moreover, as the batch BO algorithms like BOSS and BOHB sample a batch of configurations without returned evaluation results when generating new bracket, these configurations will usually be centralized in the hyperparameter space.
This phenomenon is not good for exploration. To address this issue, the Constant Liar algorithm (\cite{constant_liar}) uses a greedy approach to iteratively construct data points (the “liar”) and updates the Gaussian Process model with this “liar” value. In this way, the diversity of the sampled configurations will be increased. This technique is used to update the Gaussian process model of parallel BOSS.

\section{Experimental Results}\label{sec:app}
This section illustrates the benefits of SS and AMSS over SH and ASHA respectively by synthetic experiments. Then, we apply the proposed BOSS to a variety of machine learning problems ranging from Neural Architecture Search (NAS), Data Augmentation (DA), Object Detection (OD) to Reinforcement Learning (RL) which can all be cast as HPO problems. Their common point is that it takes a long time to evaluate a single hyperparameter configuration. 

From the results, we can see that BOSS is consistently better than BOHB, but the performance difference is little. Thus, a paired t-test for BOSS and BOHB with all the data of NAS, data augmentation, and objective detection is given. The alternative hypothesis is that BOSS is greater than BOHB. The p-value of this test is $0.03729$ which tells us the advantage of BOSS is statistically significant at $95\%$ confidence level. 

Further, since the utility of GPUs is not effective in both BOSS and BOHB, we compare the difference of GPUs' usage between BOSS and Parallel BOSS.
\subsection{Synthetic Experiments}\label{sec:sim}
In this subsection, we first compare the proposed SS and SH in synthetic experiments from two aspects, the cumulative regret and the probability of finding the true optimal configuration. 
Suppose there are $K$ configurations, and the response from the $k$-th configuration, follows the normal distribution $N(\mu_k,\sigma)$, where $\mu_k=k/K$, $\sigma$ is its standard deviation and $k\in\{0,1,...,K-1\}$. The $t$-th evaluation of the $k$-th configuration with budget $b$ means randomly sampling $b$ samples from $N(\mu_k,\sigma)$ and returning their mean denoted by $Y_t^{(k)}$. A sequence of configurations and the number of budgets used to evaluate these configurations are designed by the HPO algorithms. 


\begin{figure*}[ht]
	\begin{center}
		\includegraphics[width=0.3\textwidth]{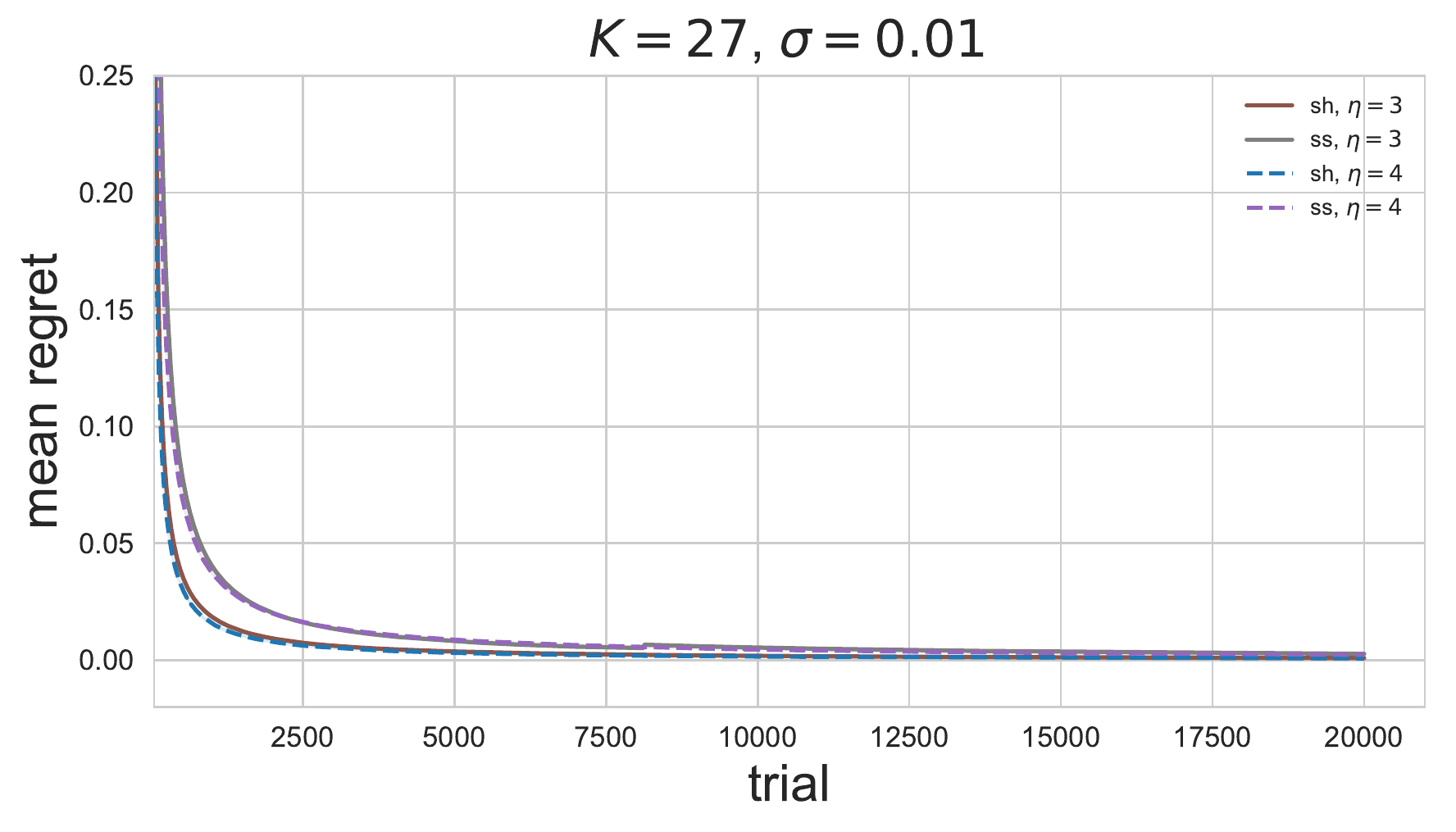}
		\includegraphics[width=0.3\textwidth]{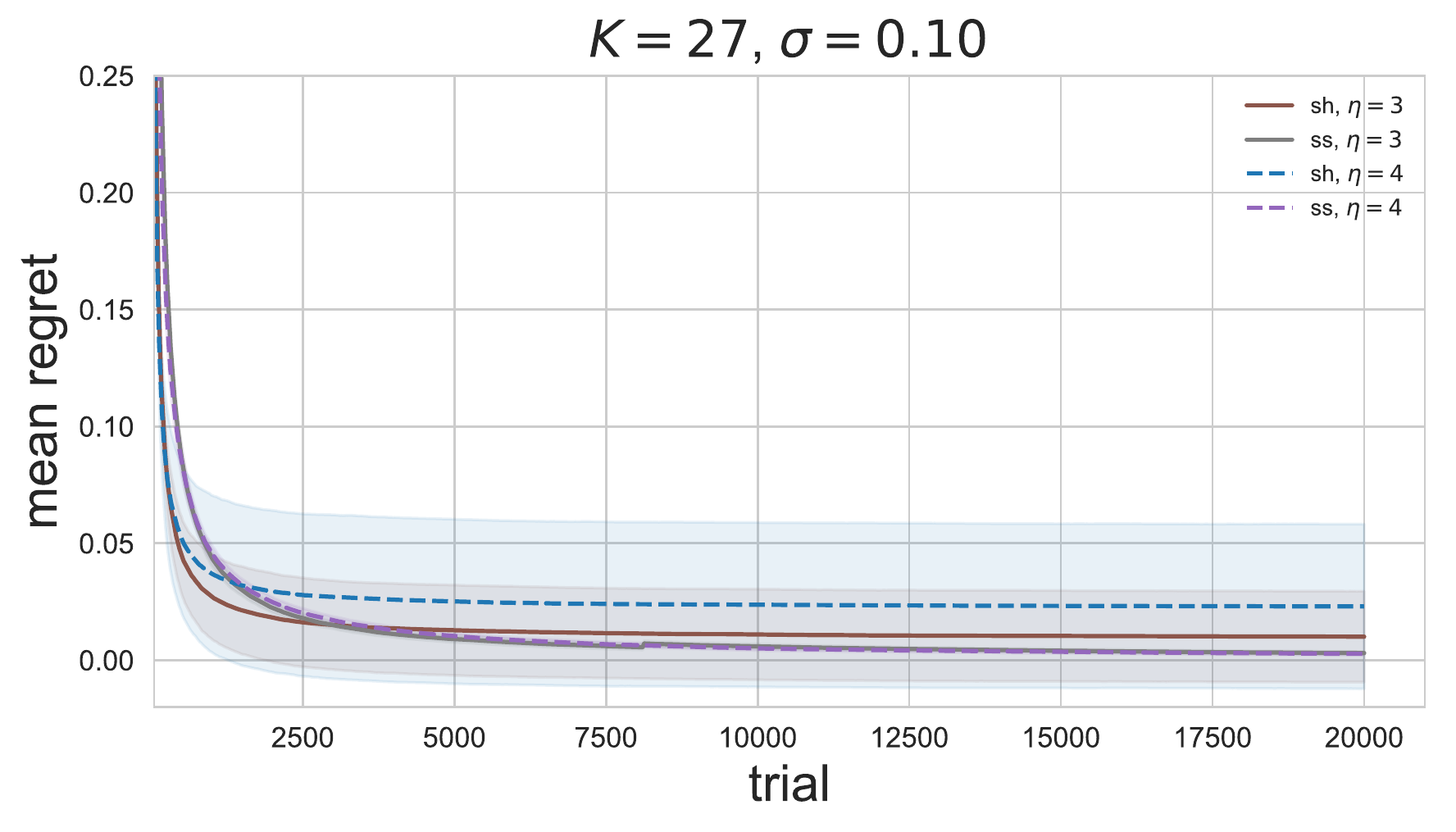}
		\includegraphics[width=0.3\textwidth]{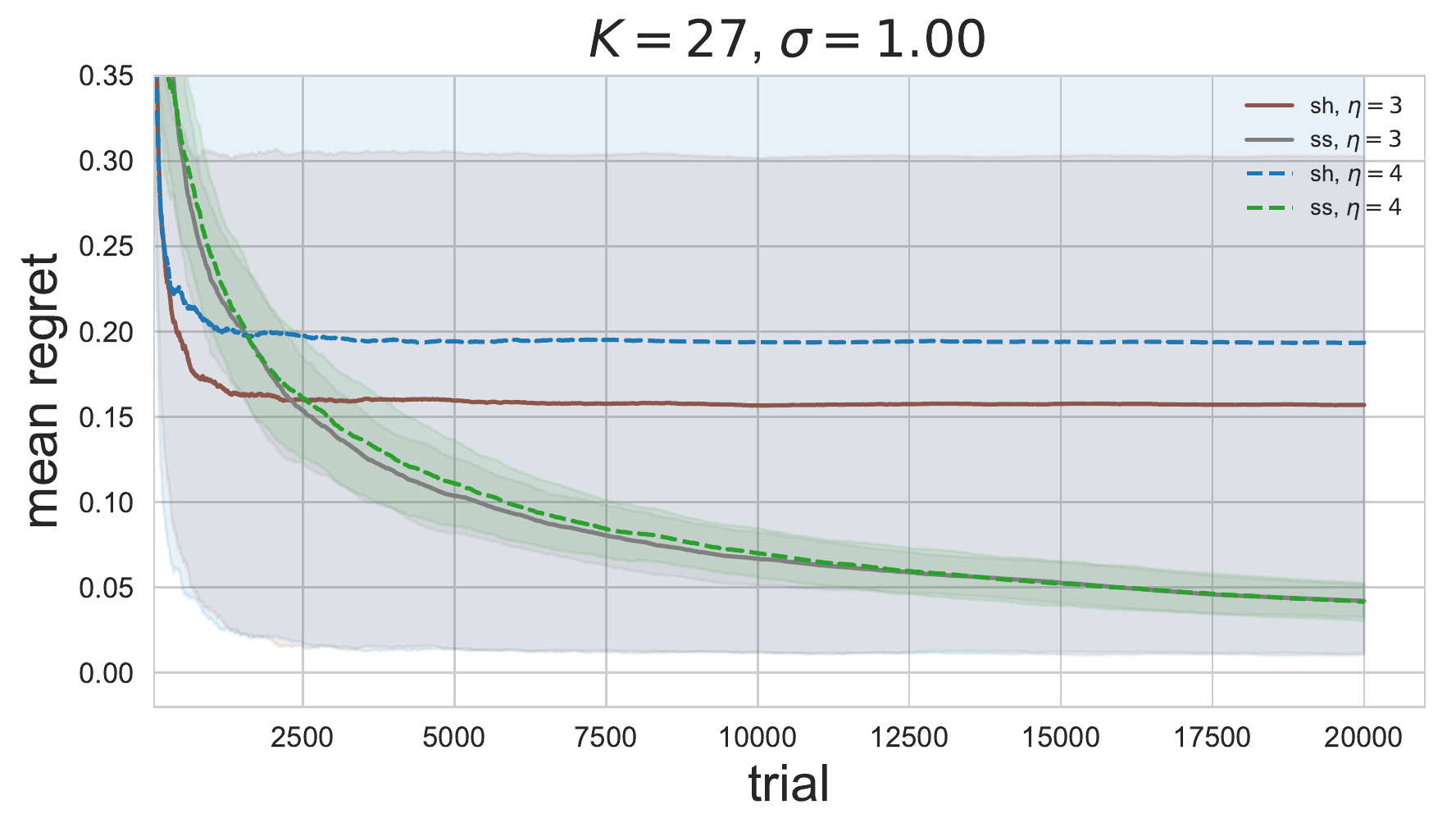}
		\caption{The comparison of the average regrets of SH and SS with different $\eta$. 
			For each setting, we conduct the experiment $50$ runs.
			The curves present the mean of the average regrets. The shaded part indicates the min/max ranges.
		}
		\label{fig:mean_regret}
	\end{center}
\end{figure*}

\begin{figure*}[ht]
	\begin{center}
		\includegraphics[width=0.32\textwidth]{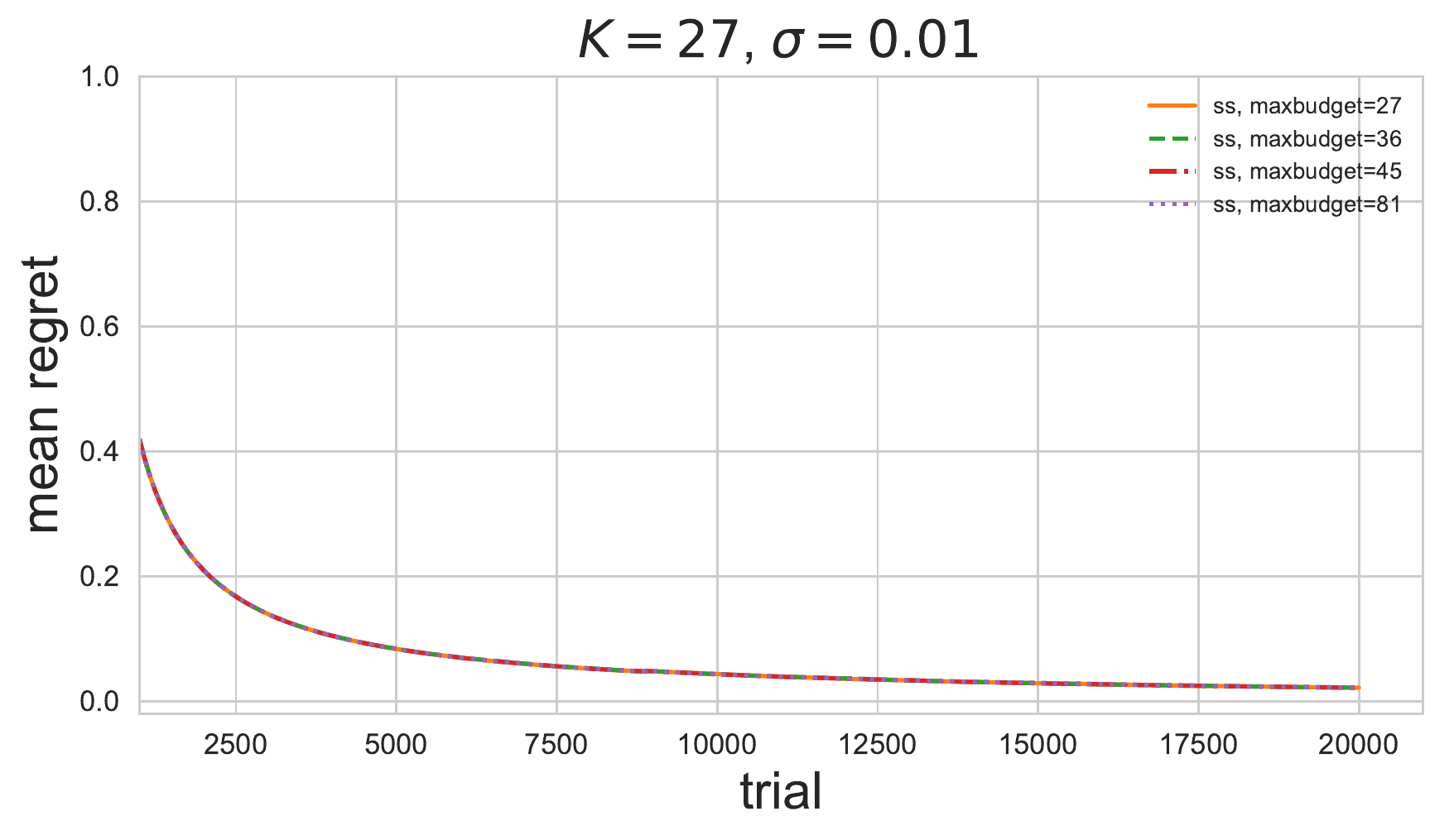}
		\includegraphics[width=0.32\textwidth]{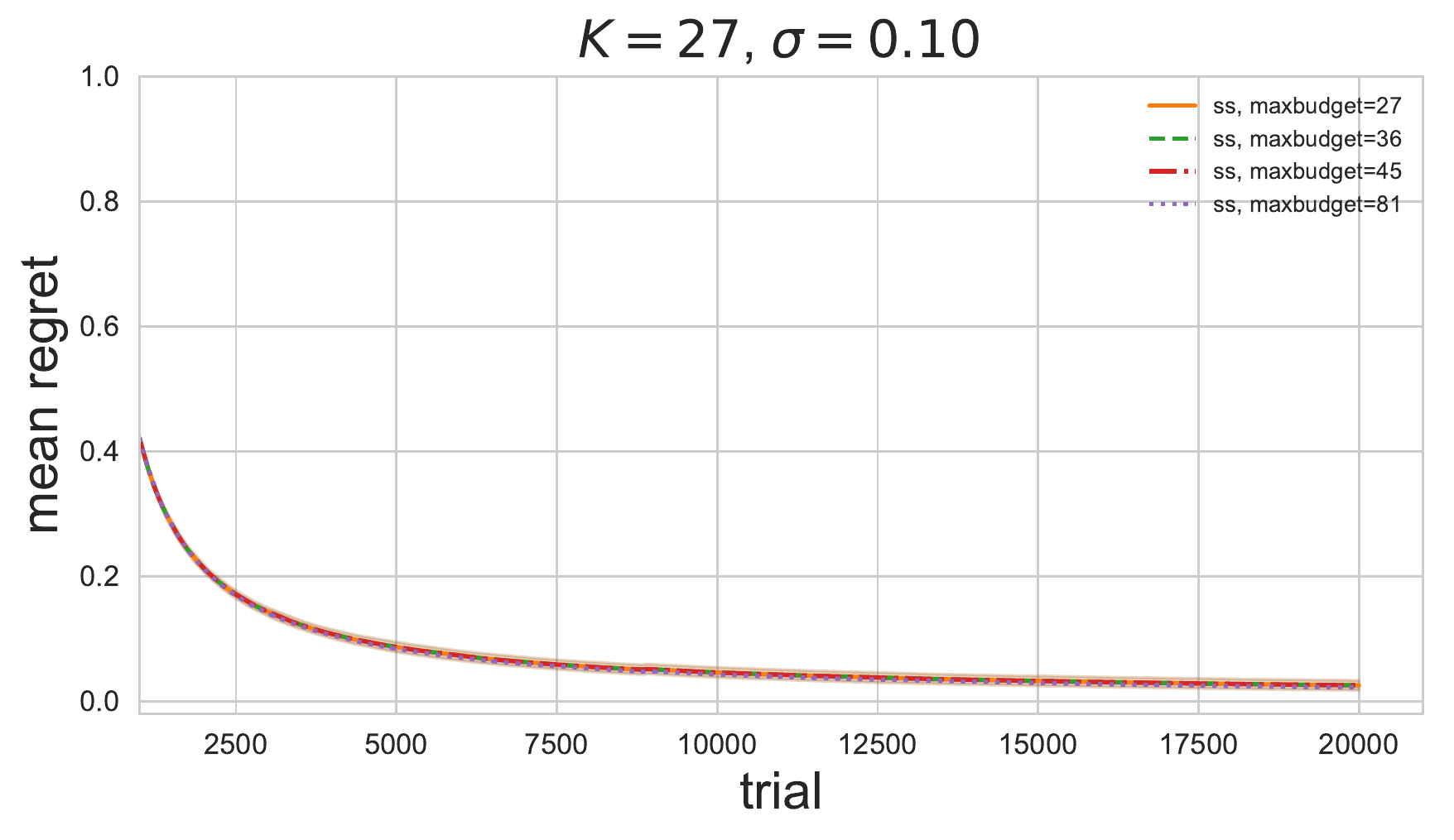}
		\includegraphics[width=0.32\textwidth]{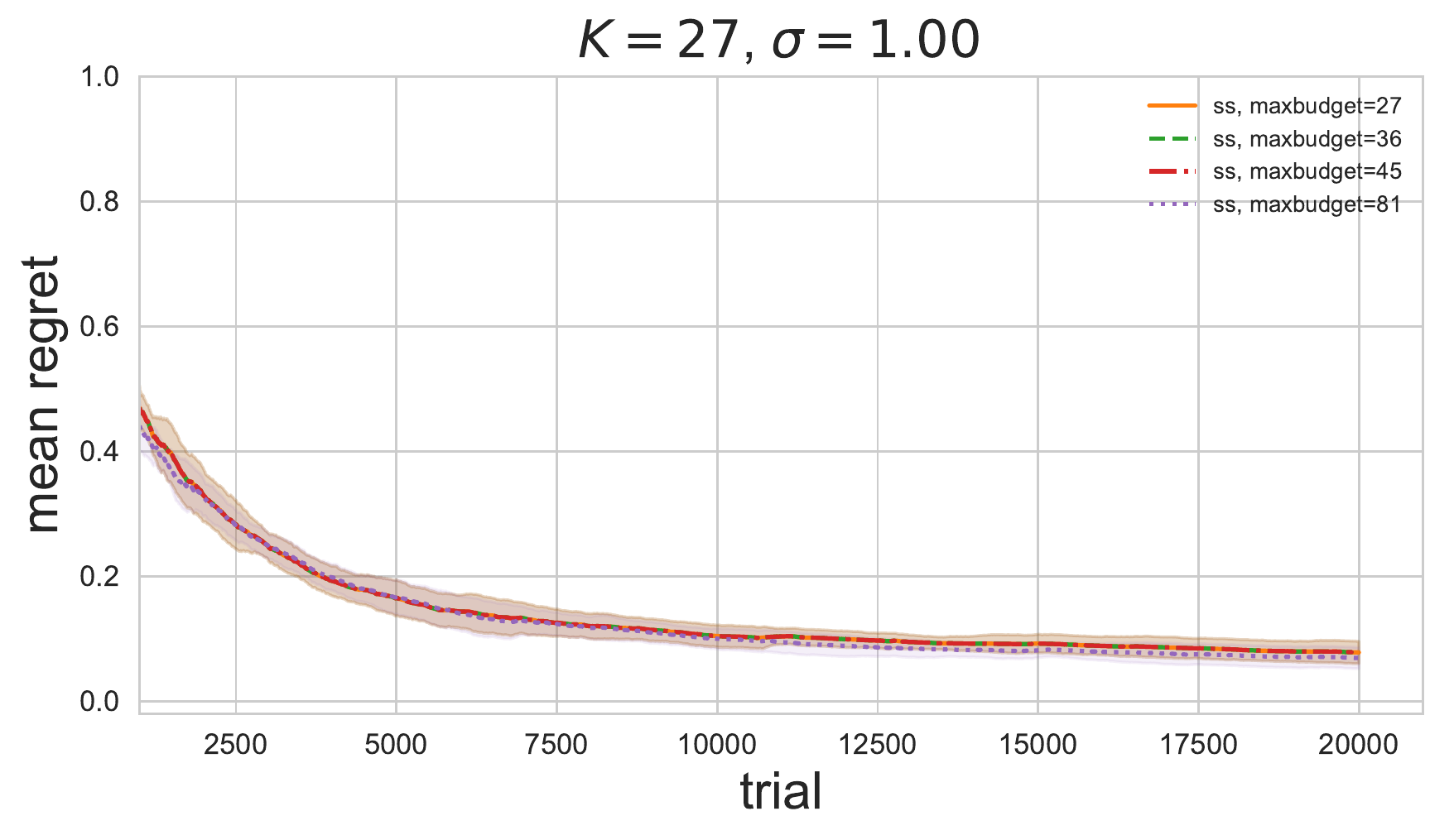}
		\caption{The comparison of the average regrets of SH and SS with different maximum budgets.
			For each setting, we conduct the experiment $50$ runs.
			The curves present the mean of the average regrets. The shaded part indicates the min/max ranges.
		}
		\label{fig:mean_regret2}
	\end{center}
\end{figure*}

Figure \ref{fig:mean_regret} demonstrates that the average regrets, defined by $ \frac{1}{N} \sum_{i=1}^N (y_i - \min_k\mu_k)$, of these algorithms are large in the beginning, since they try all configurations in the early iterations.
After that, they use collected information to determine exploration or exploitation.
For configurations with small variances, the difference between them can be easily distinguished with a small amount of budgets. Hence, SH quickly converges while SS as a more conservative policy converges slower. 
However, the performance of SH is greatly affected by the responses' instability. 
For the responses with large variances, SH fails to find the optimal configuration since it wrongly evaluates the configurations with small budgets and has no chance to fix it. On the contrary, SS is more stable to find the optimal one. For different choices of $\eta$, this conclusion still holds.
These figures also reveal that SS has more robust results than SH with different $\eta$.    
\begin{table}[h]
	\begin{center}
		\caption{The accuracy (\%) of selecting the optimal configuration.}
		\label{tab:SS_acc}
		\setlength{\tabcolsep}{1.8mm}{
			\begin{tabular}{ccccccc}
				\hline
				\multirow{2}{*}{Method} & \multicolumn{3}{c}{$K=27$}           & \multicolumn{3}{c}{$K=54$}           \\ \cline{2-7} 
				& $\sigma=0.01$ & $0.10$ & $1.00$ & $0.01$ & $0.10$ & $1.00$ \\ \hline
				SH                      & 100             & 76              & 24              & 100              & 62               & 18               \\ \hline                        
				SS                      & 100             & 100             & 100             & 100              & 100              & 88              \\ \hline
		\end{tabular}}
	\end{center}
\end{table}
Moreover, Table \ref{tab:SS_acc} demonstrates the proportion of selecting the optimal configuration by SH and SS with $\eta=3$.
Under different settings, SS performs better than SH.
The deviation of responses has a great influence on SH. 
SH gets a high accuracy in the circumstance of small deviations, but the accuracy decreases quickly as the deviation gets larger. 
When $K=27$, SS can always find the optimal configuration under different deviations even with a large deviation of $1.00$.
Note that we have compressed the mean value between $0$ and $1$, and the deviation of $1.00$ can make it difficult to distinguish between configurations. 
When $K=54$, it sometimes fails to get the optimal configuration since it needs more steps to do exploration and exploitation.
We also consider the effect of maximum budgets. Figure \ref{fig:mean_regret2} shows the results with $\eta=3$ and different maximum budgets. The performances of the algorithms are highly similar. The parameter $q_n$ equals $\sqrt{\log n}$.

\begin{figure*}[ht]
	\begin{center}
		\includegraphics[width=0.32\textwidth]{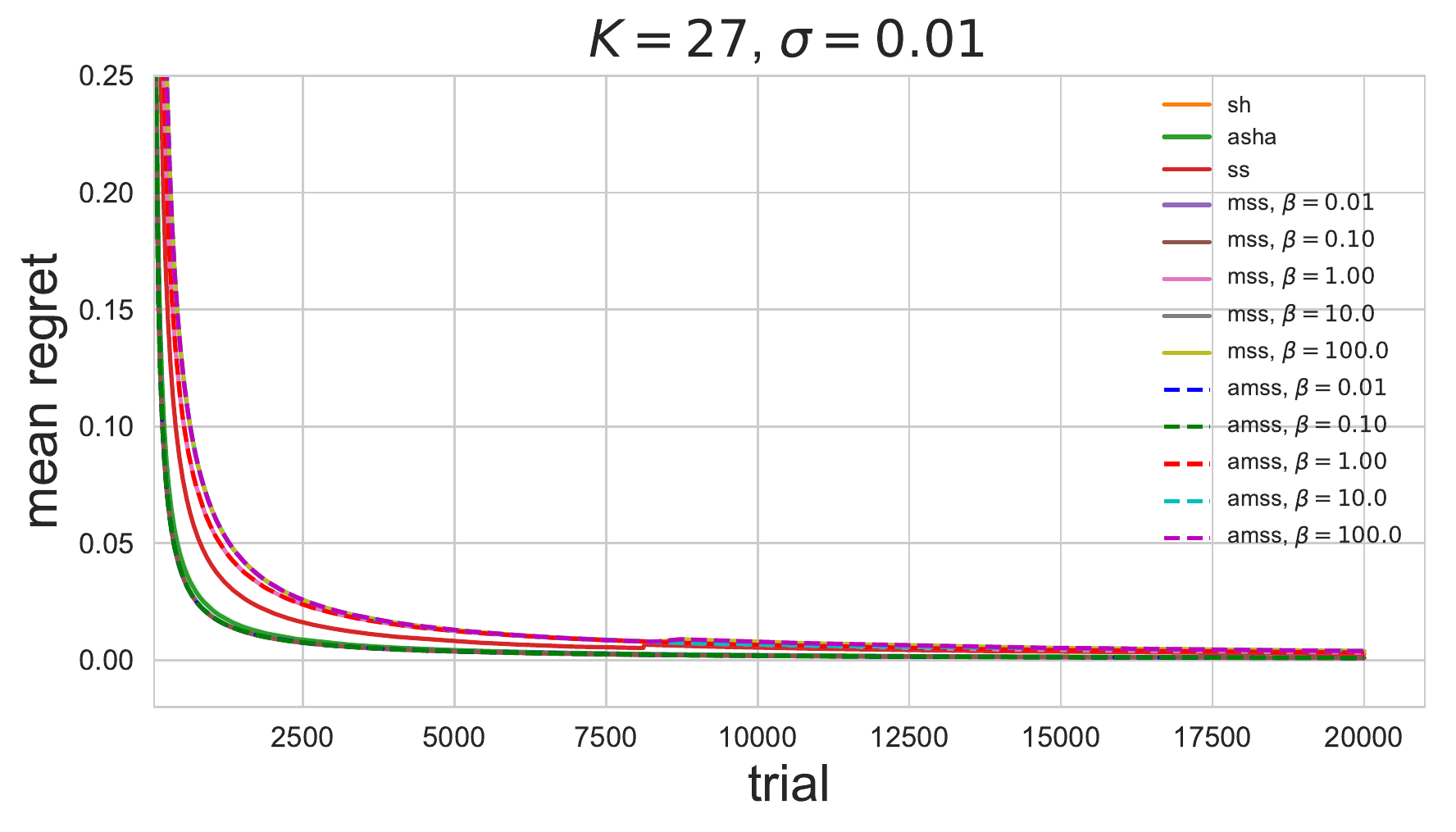}
		\includegraphics[width=0.32\textwidth]{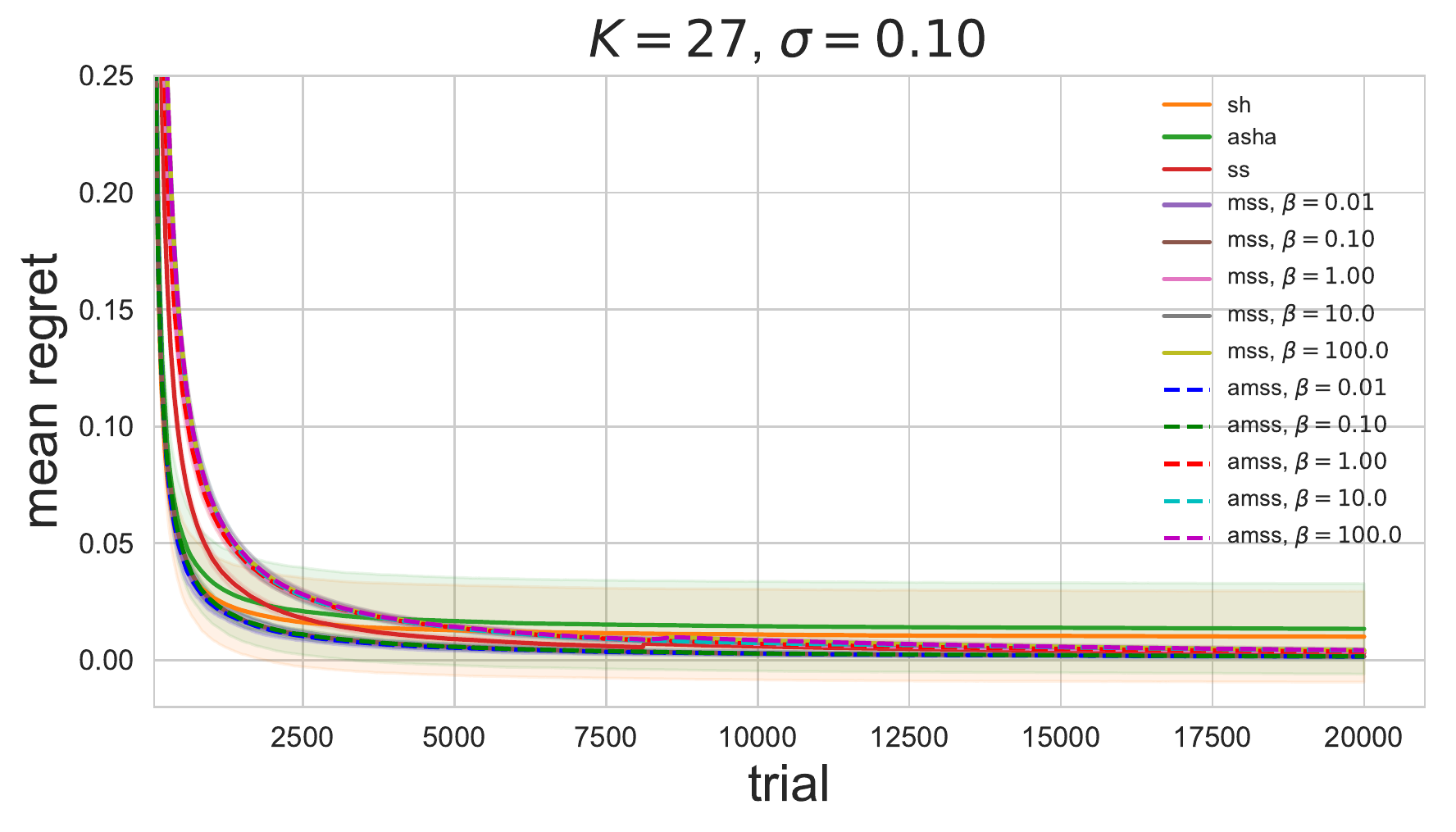}
		\includegraphics[width=0.32\textwidth]{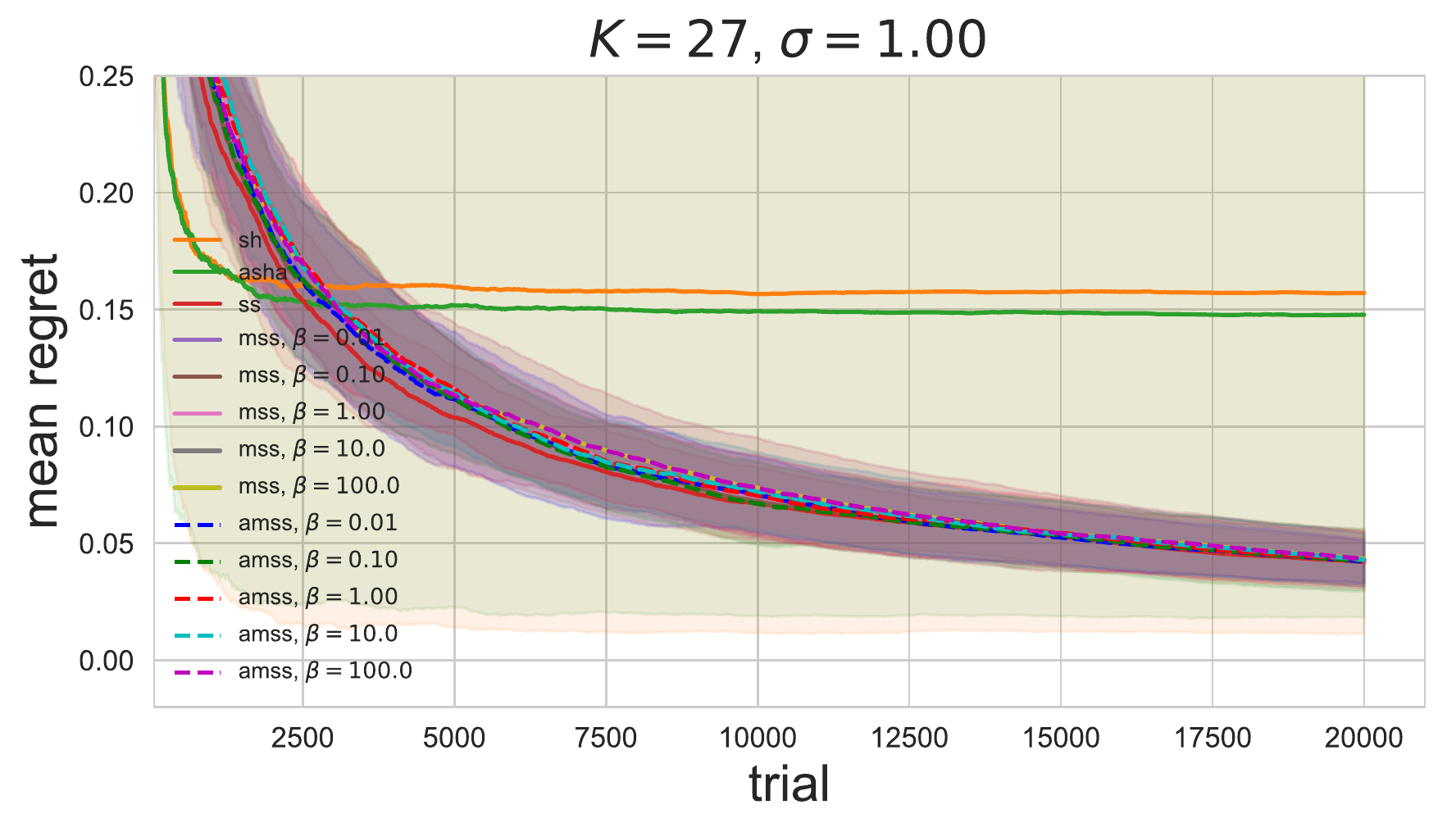}
		\caption{The comparison of the average regrets of SH, ASHA, SS, MSS and AMSS. 
			For each setting, we conduct the experiment $50$ runs.
			The curves present the mean of the average regrets. The shaded part indicates the min/max ranges.
		}
		\label{fig:mean_regret3}
	\end{center}
\end{figure*}
Figure \ref{fig:mean_regret3} illustrates the comparison of SH, ASHA, SS, MSS and AMSS. For the new parameter $\beta$, we also make a comparison to find its effect.

We can see that MSS and AMSS outperform SH and ASHA in all circumstances. 
In the cases with small deviations, these algorithms converge to the same point. 
In the cases with large deviations, the advantages of MSS and AMSS over SH and ASHA are more obviously. As for the effect of the parameter $\beta$, it reveals that $\beta$ can control the degree of conservation. The algorithms with large $\beta$ take conservative behaviors to explore more configurations in the initial stages.  
This phenomenon is especially obvious in the context of small deviations.
When $\sigma=0.01$, the performances of $\beta=100$ are the worst at the beginning.
The reason is that large $\beta$ considers the number of observations only. They ignores the current performance of the configurations. However, their final performances are highly similar. Note that AMSS has the same result with the same trials, and asynchronous parallelization can reduce running time which makes AMSS more efficient.

\subsection{Data Augmentation Search}\label{sec:aa}
Data augmentation is an effective technique to generate more samples from data by rotating, inverting or other operations for improving the accuracy of image classifiers. However, most implementations are manually designed with a few exceptions. \citet{cubuk2018autoaugment} proposed a simple procedure called AutoAugment to automatically search for improved data augmentation policies. Unfortunately, it is very time-consuming, e.g., it takes $5000$ GPU hours in searching procedure for CIFAR100 \citep{krizhevsky2009learning}. More recently, \citet{ho2019population} and \citet{lim2019fast} designed more efficient algorithms for this particular task. 

In their search space, a policy consists of $5$ sub-policies with each sub-policy consisting of
two image operations to be applied in sequence. Additionally, each operation is also associated with two hyperparameters: (1) the probability of applying the operation, and (2) the magnitude of the operation. In total, there are $16$ operations in the search space. Each operation also comes
with a default range of magnitudes. These settings are described in \citet{cubuk2018autoaugment}. For this problem, we need to tune two hyperparameters of each sub-policy and choose the best five sub-policies to form a policy. This is a natural HPO problem. BOSS can be adopted directly without any modification. As for the setting of the parameters $\eta$ and $R$ in Algorithms \ref{alg:BOSS}, we set the ratio $\eta=3$ refers to the same setting as SH and BOHB. The maximum budget $R$ equals to one third of the number of epochs for convergence. In our experience given in Figure \ref{fig:mean_regret2}, small changes of $R$ have no effect on the performance. The following applications all use this setting.

We search the data augmentation policy in the image classification tasks of CIFAR-10 and CIFAR-100.
We follow the setting in AutoAugment \citep{cubuk2018autoaugment} to search for the best policy on a smaller data set, which consists of $4,000$ randomly chosen examples, to save time for training child models during the augmentation search process.
For the child model architecture, we use WideResNet-28-10 (28 layers - widening factor of 10) \citep{zagoruyko2016wide}.
The augmentation policy is combined with standard data pre-processing: 
on one image, we normalize the data in the following order, use the horizontal flips with $50\%$ probability, zero-padding and random crops, augmentation policy, and finally Cutout with $16\times16$ pixels \citep{devries2017improved}.
We run BOSS, BOHB and HB with budgets of $1, 5, 16, 50, 150$ epochs using $32$ parallel workers for $10$ iterations. 
We run BO, SH and Random Search with budgets of $150$ epochs using $32$ parallel workers for $10$ iterations. 
Every two workers use one NVIDIA V100 GPU for parallel training.
For each sub-task, we use a SGD optimizer with a weight decay of $0.0005$, momentum of $0.9$, learning rate of $0.1$.

We use the found policies to train final models on CIFAR-10, CIFAR-100 with $200$ epochs.
All the results of the baselines are replicated in our experiments and match the previously reported results \citep{falkner2018bohb,cubuk2018autoaugment}.
Instead of searching the augmentation policy in a large amount of GPU days, we use the searched policy reported in \citet{cubuk2018autoaugment} to evaluate its performance.

\begin{figure}[ht]
	\begin{center}
		\includegraphics[width=0.35\textwidth]{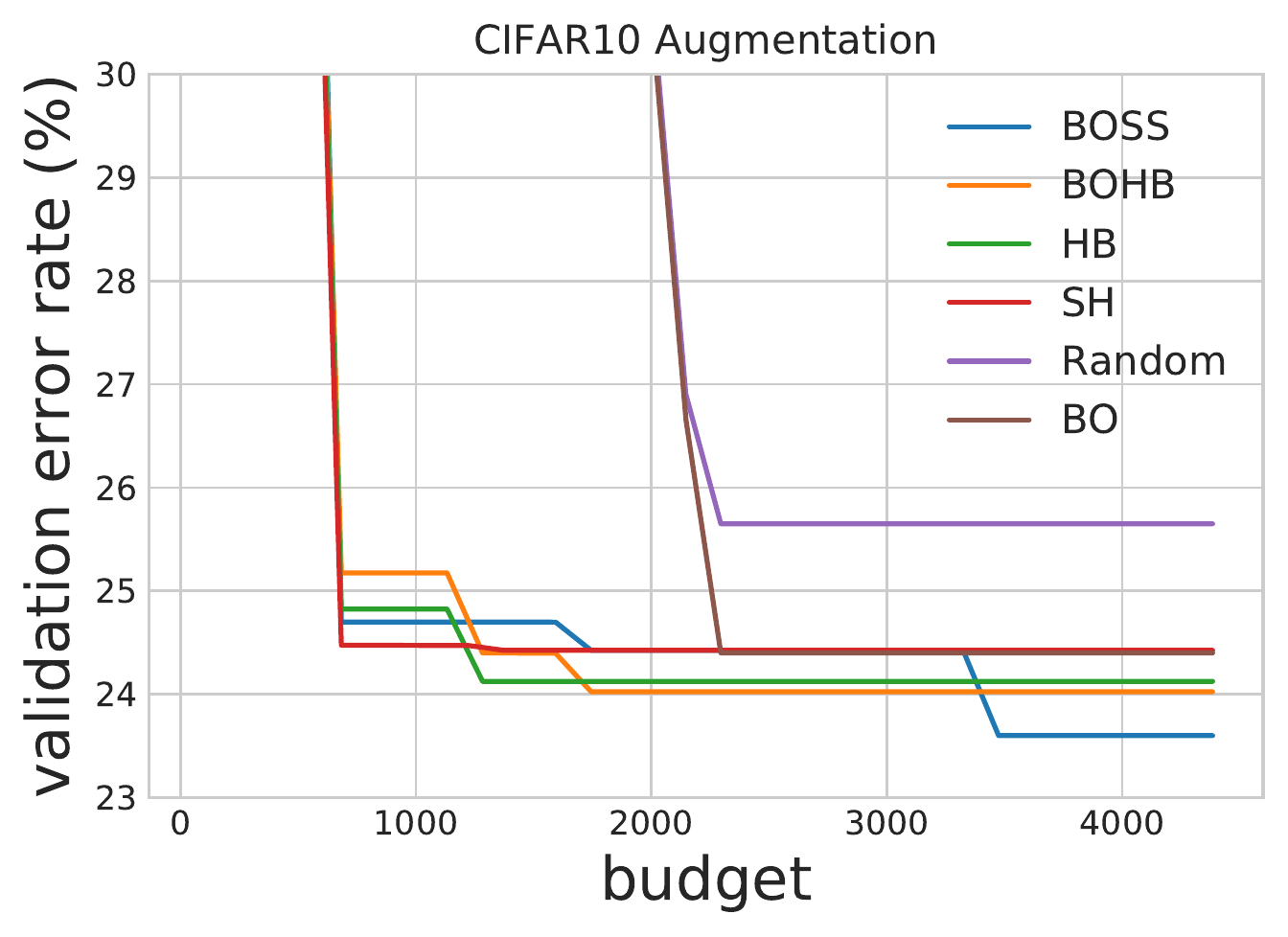}
		\includegraphics[width=0.35\textwidth]{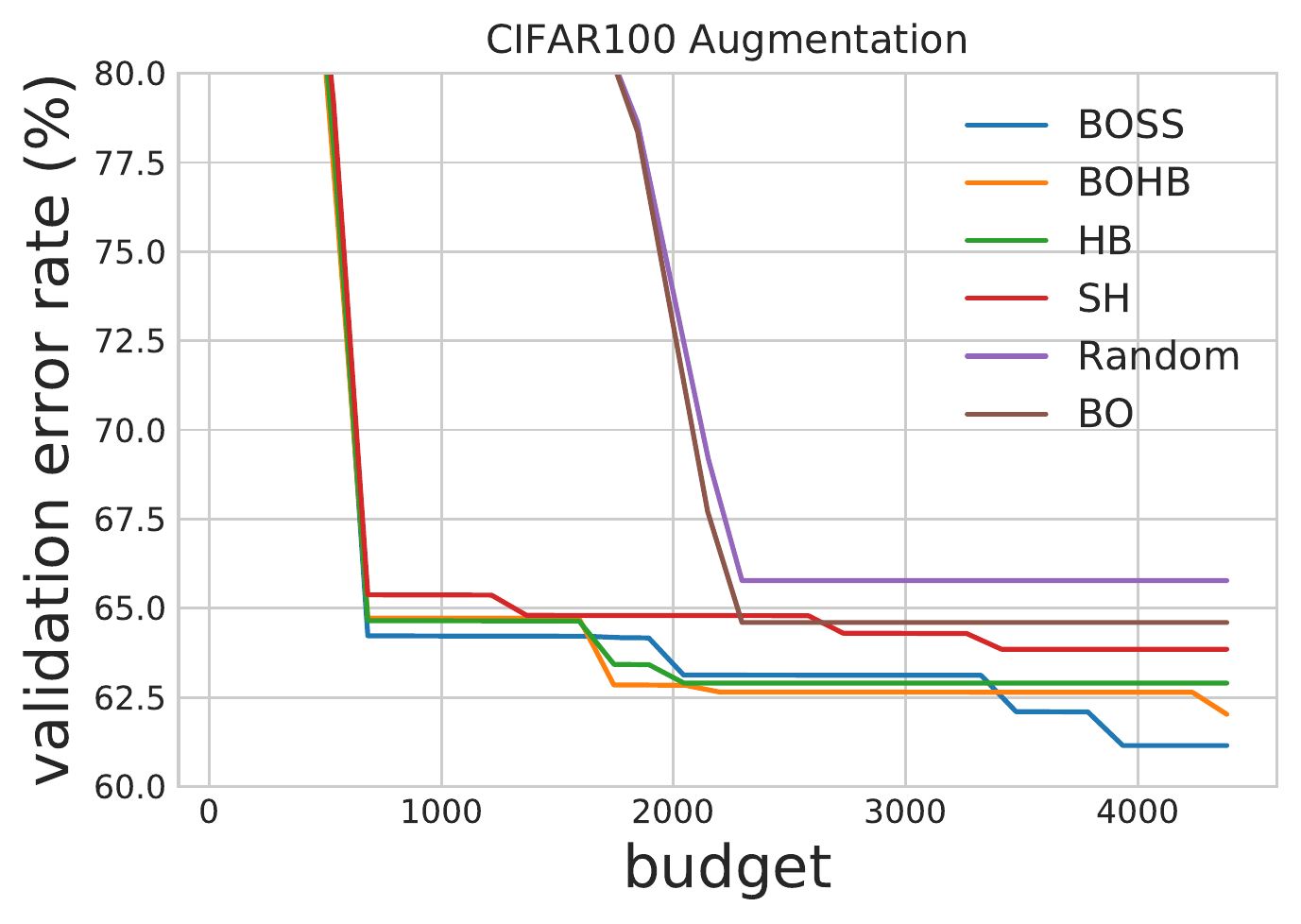}
		\caption{Data Augmentation (DA) as a hyperparameter optimization (HPO) problem. 
		}
		\label{fig:traningAA}
	\end{center}
\end{figure}

Figure \ref{fig:traningAA} demonstrates the performance during the searching procedure.
The budget means the total epochs used in the searching process. 
The results show that BOHB performs better than BOSS in the beginning, but BOSS converges to better configurations in the end. This is caused by the conservative and the asymptotic optimality of BOSS.
The performances of BO, HB, SH and Random Search are weaker than BOHB and BOSS, since they either sample configurations uniformly without considering the information brought by previous trials or evaluate configurations slowly.
Note that the error rates of these methods are high especially on CIFAR100. This is because in the searching procedure, used budgets are much smaller than the training procedure for the searched policy.

The results listed in Table \ref{tab:data_augmentation} show that the naive HPO methods including Random Search, SH, HB and BO have similar results as the original RL-based DA method, and the proposed efficient search scheme BOSS has the best accuracy and BOHB is close behind.

\begin{table}[ht]
	\begin{center}
		\caption{Top-1 accuracy (\%) of data augmentation on the test dataset of CIFAR-10 and CIFAR-100. All of the reported results are averaged over 5 runs.}
		\label{tab:data_augmentation}
		\setlength{\tabcolsep}{4.8mm}{
		\begin{tabular}{ccc}
			\hline
			\multirow{2}{*}{Method} & \multicolumn{2}{c}{Test Accuracy (std)}   \\ \cline{2-3}
			& CIFAR10                 & CIFAR100                            \\ \hline
			AA                      & 97.09 (0.14)                  & 82.42   (0.17)                                                                                                    \\ \hline
			Random                  & 97.01 (0.11)                    & 82.41 (0.23)                                                                                                       \\ \hline
			SH                      & 96.93 (0.14)                   & 81.49 (0.23)                                                                                                       \\ \hline
			HB                      & 97.00 (0.10)                   & 82.07 (0.11)                                                                                                       \\ \hline
			BO                      & 97.11 (0.16)                   & 82.19  (0.25)                                                                                                     \\ \hline
			BOHB                    & 97.25 (0.13)          & 82.52 (0.15)                                                                                                       \\ \hline
			BOSS                    & \textbf{97.32} (0.12)                    & \textbf{83.03}  (0.22)                                                                                            \\ \hline
		\end{tabular}}		
	\end{center}
\end{table}

\subsection{Neural Architecture Search}\label{sec:nas}

One crucial aspect of the deep learning development is novel neural architectures. Designing architectures manually is a time-consuming and error-prone process. Because of this, there is a growing interest in automated neural architecture search. \citet{elsken2019neural} provided an overview of existing work in this field of research. 
We use the search space of DARTS \citep{liu2018darts} as an example to illustrate HPO methods on NAS. 
Particularly, their goal is to search for a cell as a basic unit. In each cell, there are $N$ nodes forming a fixed directed acyclic graph (DAG). Each edge of the DAG represents an operation, such as skip-connection, convolution, max pooling, etc., weighted by the architecture parameter $\alpha$. 
For the search procedure, the training loss and validation loss are denoted by $L_{train}$ and $L_{val}$ respectively. 
Then the architecture parameters are learned with the following bi-level optimization problem:
\begin{align*}
&\min_\alpha \quad L_{val}(\omega^*(\alpha),\alpha),\\
&\ \text{s.t.} \ \omega^*(\alpha)=\arg\min_\omega L_{train}(\omega,\alpha).
\end{align*}
Here, $\alpha$ are hyperparameters in the HPO framework. For evaluating $\alpha$, we need to optimize its network parameters $\omega$. It is usually time-consuming. Note that BOSS is exactly proposed to make a fast and efficient evaluation for this problem. 

We search the neural architectures in the image classification tasks of CIFAR10 and CIFAR100.
We follow the settings of DARTS \citep{liu2018darts}.
The architecture parameter $\alpha$ determines two kinds of basic units: normal cell and reduction cell. 
We run BOHB, BOSS and HB with maximum budgets of $50$ epochs using $32$ parallel workers for $10$ iterations.
We run BO, \textsl{SH} and Random Search with total budgets as the same as BOSS using $32$ parallel workers for $10$ iterations.
For a sampled architecture parameter, we fix it in the training process of updating model parameters. 
A SGD optimizer is used with learning rate of $0.025$, momentum of $0.9$, weight decay of $0.0003$, and a cosine learning decay with an annealing cycle.

\begin{figure}[ht]
	\begin{center}
		\includegraphics[width=0.35\textwidth]{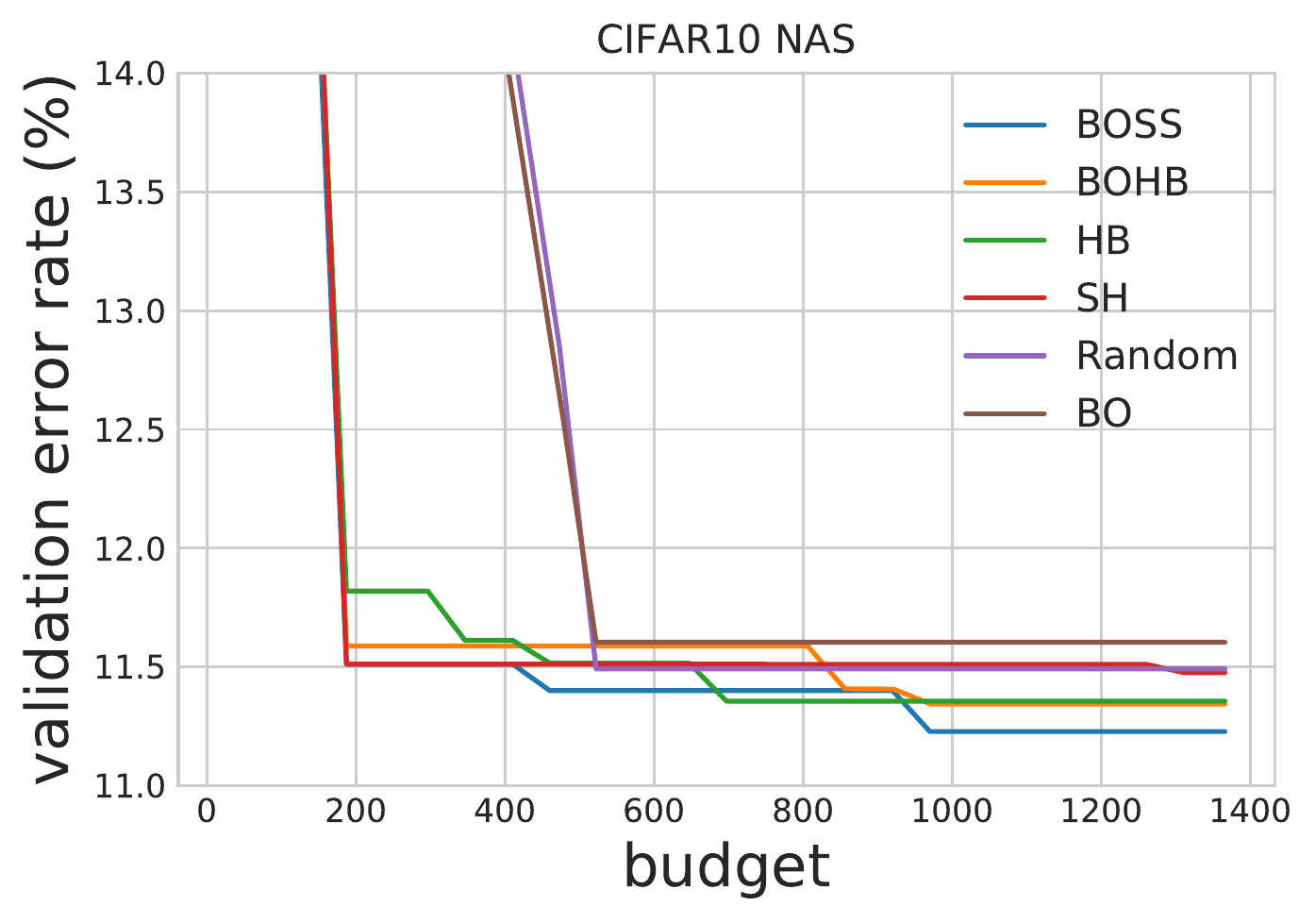}
		\includegraphics[width=0.35\textwidth]{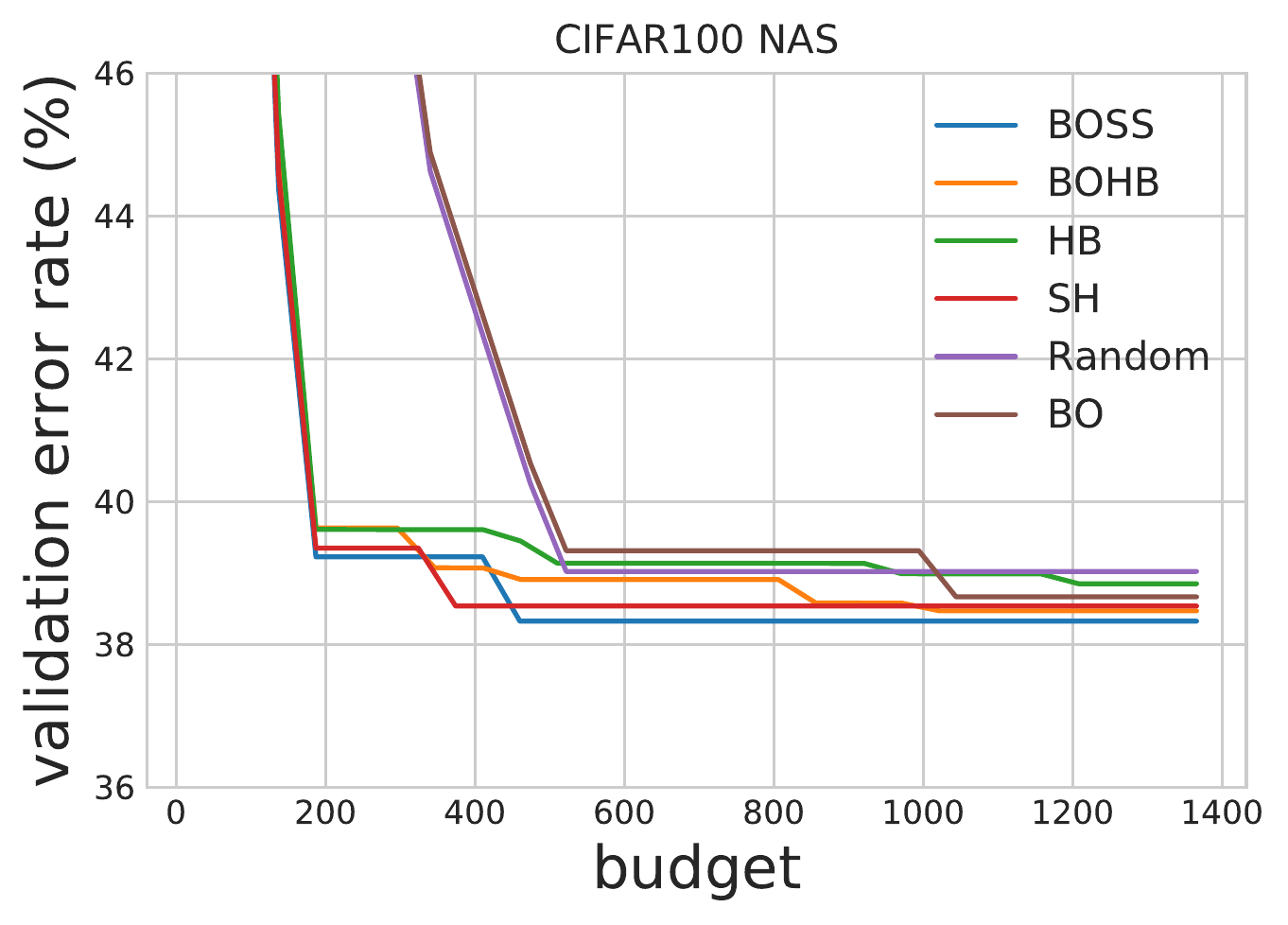}
		\caption{Neural architecture search (NAS) as a hyperparameter optimization (HPO) problem. 
		}
		\label{fig:trainNAS}
	\end{center}
\end{figure}

After searching out the optimal configuration, we evaluate it with $200$ epochs. 
Figure \ref{fig:trainNAS} shows the highly similar results of that in data augmentation.

The results are listed in Table \ref{tab:nas}. Note that the search space and search procedure of DARTS are designed on CIFAR10. The naive HPO methods including Random Search, SH, HB and BO are worse than DARTS while BOSS and BOHB have comparable accuracy. On CIFAR100, the advantage of these HPO methods increases, and BOSS is significantly better than other search procedures.

\begin{figure}[ht]
	\begin{center}
		\includegraphics[width=6.1cm, height=2.9cm]{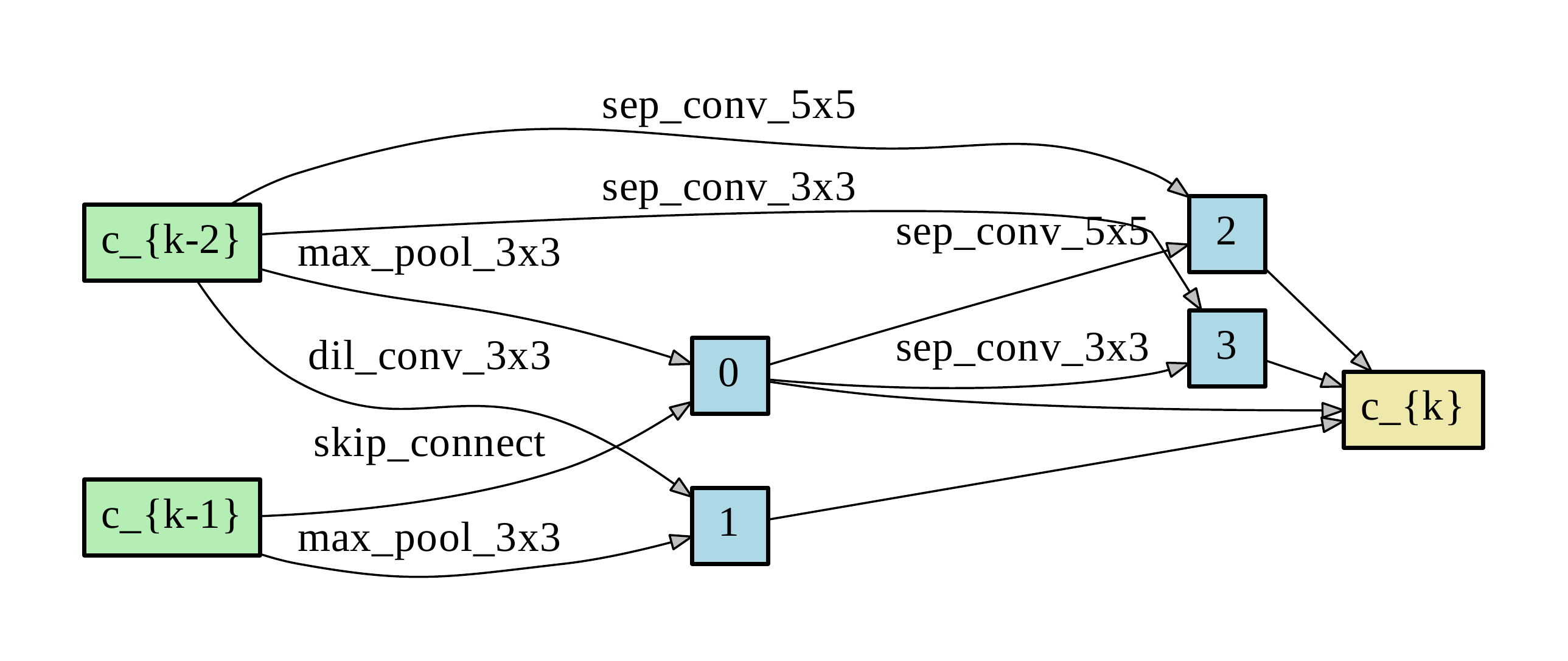}
		\includegraphics[width=7.6cm, height=2.6cm]{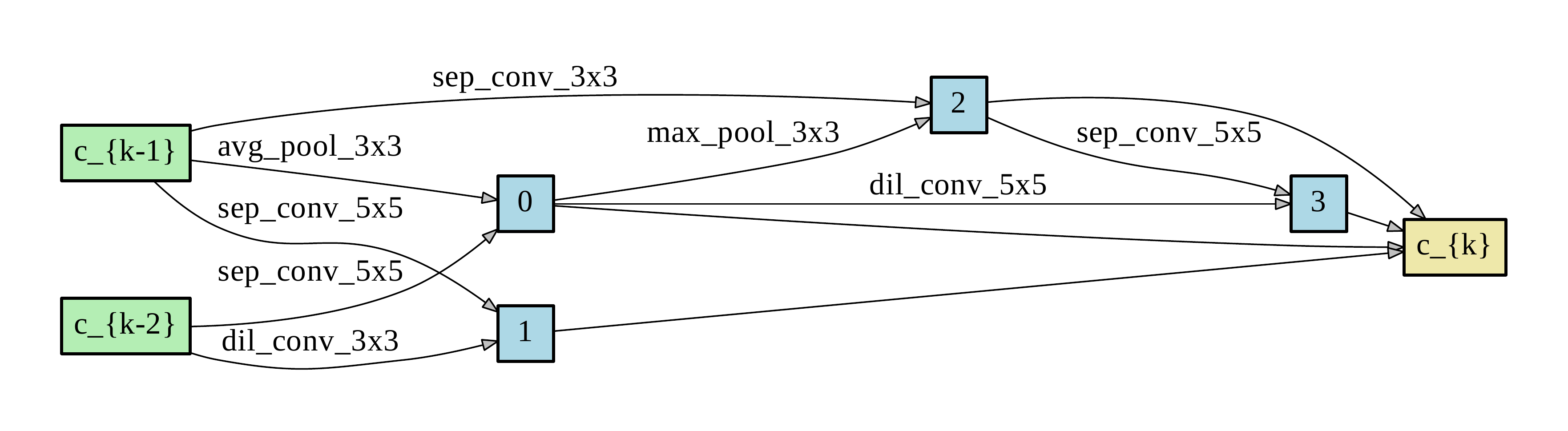}
		\caption{The architectures of normal cell (left) and reduction cell (right) learned by BOSS on CIFAR10.}
		\label{fig:skip10}
	\end{center}
\end{figure}

\begin{figure}[ht]
	\begin{center}
		\includegraphics[width=6.1cm, height=2.6cm]{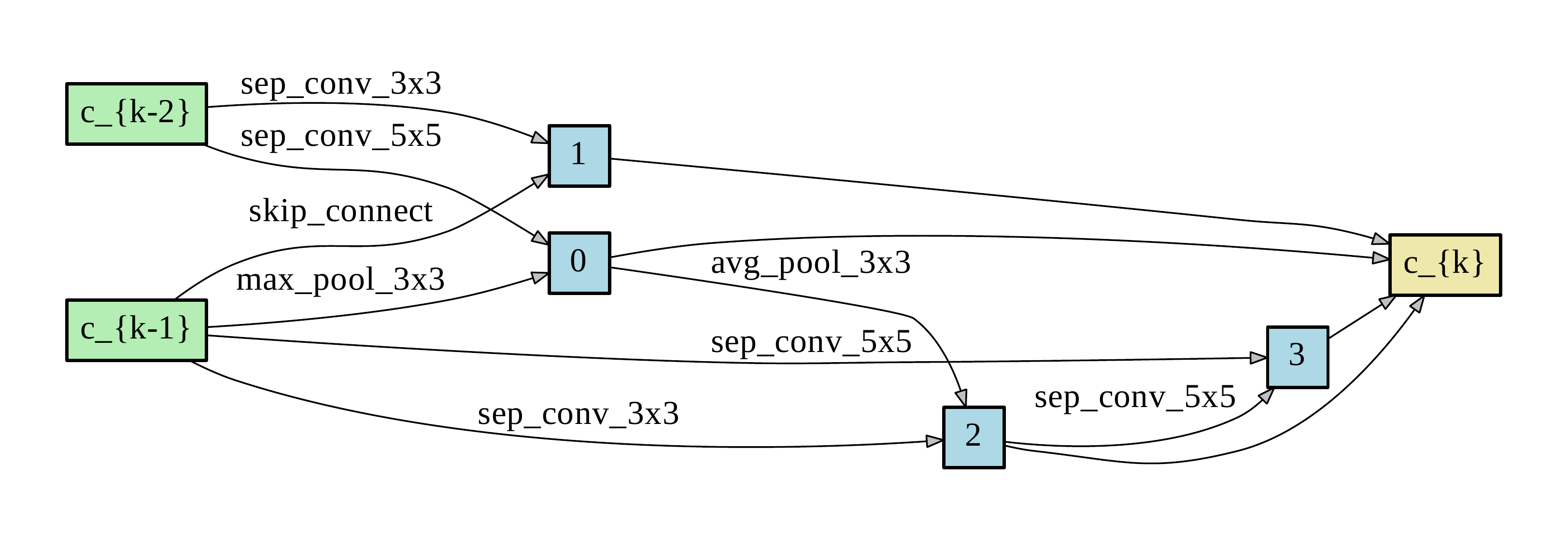}		
		\includegraphics[width=7.2cm, height=1.9cm]{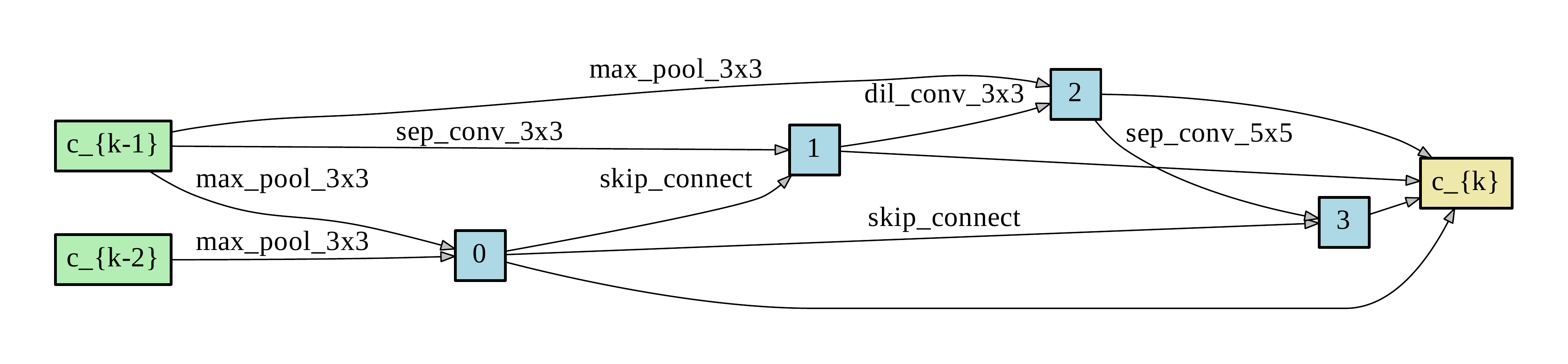}
		\caption{The architectures of normal cell (left) and reduction cell (right) learned by BOSS on CIFAR100.}
		\label{fig:skip100}
	\end{center}
\end{figure}

\begin{table}[ht]
	\begin{center}
		\caption{Test accuracy (\%) of NAS on CIFAR-10 and CIFAR-100. All of the reported results are averaged over 5 runs.}
		\label{tab:nas}
		\setlength{\tabcolsep}{4.8mm}{
		\begin{tabular}{cccc}
			\hline
			\multirow{2}{*}{Method} & \multicolumn{2}{c}{Test Accuracy (std)}    \\ \cline{2-3}
			&CIFAR10         & CIFAR100                                                   \\ \hline
			DARTS                   & 97.26 (0.13)           & 81.71  (0.31)                                                                \\ \hline
			Random                  & 96.86 (0.19)           & 81.65   (0.34)                                                                 \\ \hline
			SH                      & 96.55    (0.20)        & 81.19   (0.18)                                                                 \\ \hline
			HB                      & 96.99  (0.18)          & 81.63 (0.37)                                                                   \\ \hline
			BO                      & 96.87 (0.15)           & 82.37  (0.24)                                                                 \\ \hline
			BOHB                    & 97.23 (0.17)           & 82.67 (0.27)                                                                   \\ \hline
			BOSS                    & \textbf{97.29} (0.15)  & \textbf{83.10}  (0.23)                                                        \\ \hline
		\end{tabular}}
	\end{center}
\end{table}
Moreover, when transferred to ImageNet in the mobile setting, the same architectures searched from CIFAR10 and CIFAR100 by BOSS achieve high accuracy of $74.46\%$ and $74.82\%$ respectively while DARTS reported the accuracy of $73.3\%$. This phenomenon reveals the weakness of DARTS in that it is prone to search architectures with many skip-connect operations which is not preferred. 
\citep{zela2020understanding,liang2019darts+} reduced the number of skip-connect operations by early stopping. 
However, this issue disappears in BOSS naturally which is depicted in Figures \ref{fig:skip10} and \ref{fig:skip100}, because DARTS changes architecture parameters and network parameters in turn while BOSS does not change the architecture during the training of network.

\subsection{Object Detection}\label{sec:od}

Object detection is to use an anchor box to cover the target object. Most state-of-the-art OD systems follow an anchor-based diagram. Anchor boxes are densely proposed over the images and the network is trained to predict the boxes' position offset as well as the classification confidence. Existing systems use ad-hoc heuristic adjustments to define the anchor configurations based on predefined anchor box shapes and size. However, this might be sub-optimal or even wrong when a new dataset or a new model is adopted. Hence, the parameters of anchor boxes for object detection need to be automatically optimized including number, scales, and ratios of anchor boxes.

In the literature, the anchor shapes are typically determined by manual selection \citep{dai2016r,liu2016ssd,ren2015faster} or naive clustering methods \citep{redmon2017yolo9000}. Different from the traditional methods, there are several works focusing on utilizing anchors more effectively and efficiently \citep{yang2018metaanchor,zhong2018anchor}. 
\begin{table*}[ht]
	\caption{Comparisons of different hyperparameter searching schemes with the same training procedure on MSCOCO.}
	\label{tab:od}
	\scalebox{0.8}{
		\begin{tabular}{c|c|c|c|c|c|c|c|c|c|c|c|c}
			
			\hline	
			Method& $mAP$ & $AP_{50}$& $AP_{75}$ &$AP_S$& $AP_M$ &$AP_L$& $AR_1$& $AR_{10}$ &$AR_{100}$& $AR_{S}$ &$AR_{M}$ &$AR_{L}$\\
			\hline
			Manual Search& 36.4& 58.2& 39.1& 21.3& 40.1& 46.5& 30.3& 48.8& 51.3& 32.1& 55.6& 64.6\\
			$K$-Means& 37.0& 58.9& 39.8& 21.9& 40.5& 48.5& 31.0& 49.3& 51.9& 32.9& 57.5& 66.3\\
			Random Search& 37.2& 58.8& 39.9& 21.7& 40.6& 48.1& 31.1& 49.5& 52.1& 33.8& 57.3& 66.5\\
			BOHB& 37.8& 58.9& 40.4& 22.7& 41.3& 49.9& 31.8& 50.2& 52.5& 34.2&  \textbf{57.5}& 65.2\\
			BOSS&  \textbf{38.8}&  \textbf{60.7}&  \textbf{41.6}&  \textbf{23.7}&  \textbf{42.5}&  \textbf{51.5}&  \textbf{32.3}&  \textbf{51.2}&  \textbf{54.0} & \textbf{35.5}& 57.3&  \textbf{68.2}\\
			\hline
	\end{tabular}}
\end{table*}

Table \ref{tab:od} compares BOSS with several existing anchor initialization methods as follows.
\begin{itemize}
	\setlength{\itemsep}{0pt}
	\setlength{\parsep}{0pt}
	\setlength{\parskip}{0pt}
	\item [(a)] Set anchor scales and ratios like most detection systems by manually searching.
	\item [(b)] Use $K$-means method proposed in YOLOv2 \citep{redmon2017yolo9000} to obtain clusters and treat them as initial anchors. 
	\item [(c)] Use random search to determine anchors. 
	\item [(d)] Use BOHB to determine anchors. 
	\item [(e)] Use BOSS to determine anchors. 
\end{itemize}
The standard criteria of object detection, mean Average Precision (mAP) and Average Recall (AR), are used to measure the performance. The subscript of $AP_{50,75}$ means different IoU thresholds, and the subscript of $AR_{1,10,100}$ represents different numbers of given objects per image. S, M and L refer to small, medium and large size of an object respectively. For fairness of comparing these different search schemes, we use the same training procedure which is Faster-RCNN \citep{ren2015faster} combined with FPN \citep{lin2017feature} as detector, ResNet-50 \citep{he2016deep} as backbone on MSCOCO \citep{lin2014microsoft}.

The results reveal that all three HPO methods outperform two classical ones. Moreover, BOSS has uniformly better performance than other two HPO methods which shows the usefulness and effectiveness of BOSS. In addition, since the training procedures of MetaAnchor \citep{yang2018metaanchor} and Zhong's method \citep{zhong2018anchor} are different from ours, we just compare the difference before and after using hyperparameter searching. BOSS brings about $2.4\%$ $mAP$ improvement while MetaAnchor and Zhong's method increase $mAP$ by $1.0\%$ and $1.1\%$, respectively. This advantage of BOSS is that it considers both sampling strategy and efficient evaluation while the other two methods just consider the former.

\subsection{Reinforcement Learning}\label{sec:rl}

In last few years, several different approaches have been proposed for reinforcement learning with neural network function approximators, e.g., deep Q-learning \citep{mnih2015human}, ``vanilla'' policy gradient methods \citep{mnih2016asynchronous}, trust region policy gradient methods \citep{schulman2015trust}, and proximal policy optimization (PPO) \citep{schulman2017proximal}. In these methods, there are a few hyperparameters that need to be determined. As an example, we tune the hyperparameters for the PPO Cartpole task \citep{falkner2018bohb} including the numbers of units in layers $1$ and $2$, batch size, learning rate, discount, likelihood ratio clipping, and entropy regularization. 
Different from the previous three applications, the budget becomes the number of trials used by an agent.
We run each configuration for nine trials and report the average number of episodes until the PPO has converged, which means that the learning agent achieves the highest possible reward for five consecutive episodes. 
For each configuration, we stop training after the agent has either converged or ran for a maximum of $1000$ episodes. 
For each hyperparameter optimization method, we implement five independent runs.

\begin{table}[h]
	\centering
	\caption{The epochs needed to learn a game with policies obtained by different HPO methods.}
	\label{tab:rl}
	\begin{tabular}{cccc}
		\hline
		Method & Random & BOHB & BOSS\\
		\hline
		Min terminated epoch & 102.55& 80.33& \textbf{72.77}\\
		\hline		
	\end{tabular}
\end{table}

Table \ref{tab:rl} shows that the agent using the policy obtained by BOSS is the fastest learner. The agent in RL needs to learn a stable policy which matches the conservation of BOSS.

\subsection{Parallel BOSS}
\label{sec:parallel_boss_exp}

The parallel BOSS (Algorithm~\ref{alg:parallelBOSS}) is an aggresive gpu-efficient version of BOSS.
To evaluate its performance, we implement it for a challenging industrial feature classification task with deep neural networks. The training set and testing set contain acoustic feature of around 75000 and 15000 samples. 
The feature is effectively augmented, resulting in 2.2TB training data and 11GB testing data.
The performance metric is the classification precision (higher better). Here we compare BOSS with BOHB and TPE (parallel version implemented with constant liar algorithm) algorithms under two settings provided in Table~\ref{tab:parallel_exp_setting}.

\begin{table}[h]
	\centering
	\caption{Experiment settings.}
	\label{tab:parallel_exp_setting}
	\begin{tabular}{ccccc}
		\hline
		Setting & min budget & max budget & $\eta$ & max duration\\
		\hline
		No. 1 & 1*2 & 27*2 & 3 & 60h\\
		No. 2  & 1*5 & 27*5 & 3 & 90h\\
		\hline
	\end{tabular}
\end{table}

\begin{table}[h]
	\centering
	\caption{Comparisons of different HPO algorithms on a feature classification task under setting 1 and setting 2. Due to the limited computation resource, we only compare TPE and parallel BOSS for setting 2.}
	\label{tab:parallel_feature_class_s1}
	\begin{tabular}{ccccc}
		\hline
		\multirow{2}{*}{Method} & \multicolumn{2}{c}{Setting 1}  & \multicolumn{2}{c}{Setting 2}\\
		\cline{2-5}
		& Avg top 10 & Best & Avg top 10 & Best\\
		\hline
		TPE & 0.8871 & 0.8886 & 0.8924 & 0.8936\\ 
		BOHB & 0.8803 & 0.8868  &  - & -\\
		BOSS & 0.8866 & 0.8890 & - & -\\
		parallel BOSS & \textbf{0.8908} & \textbf{0.8918}  & \textbf{0.8935} & \textbf{0.8944} \\
		\hline
	\end{tabular}
\end{table}

\begin{figure}[h!]
	\begin{center}	
		\includegraphics[height=3.5cm]{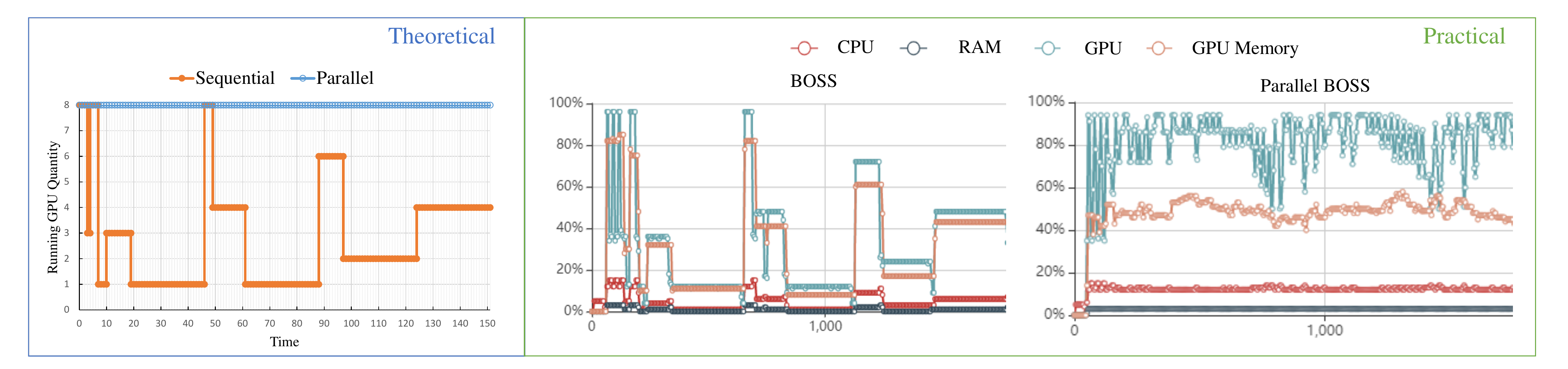}
		\caption{The usage of GPU in a complete BOSS period.}
		\label{fig:exp_gpu_use}
	\end{center}
\end{figure}

The results in Table~\ref{tab:parallel_feature_class_s1} show that the standard TPE also has a comparable performance since its usage of GPUs is efficient. 
We compare the theoretical and practical GPU usage of BOSS and parallel BOSS under setting 1, shown in Figure~\ref{fig:exp_gpu_use}. The curves of practical GPU memory useage fit well to the theoretical GPU usage.
The parallel BOSS fully utilizes the GPU resources and yields stronger results in these tasks.

\section{Conclusions}

In this paper we have proposed BOSS which combines BO and SS for HPO problems. The major contribution is to develop SS as an efficient and effective evaluation procedure and its asynchronously parallel version MSS. The proposed methods are for fast hyperparameter search evaluation by measuring the potential of hyperparameter configurations. It promises to improve robustness compared to algorithms based on successive halving. The main result is the asymptotic optimality of the cumulative regret of the proposed SS algorithm. This advantage is suitable for the BO iteration since it can collect high-quality data to estimate the surrogate model more efficiently. It can get the evaluation with less budgets which is a kind of early-stop methods.
Experiments show that SS can find the best configuration quickly and correctly and BOSS works for many popular applications. Future work to improve BOSS may further learn the search space \citep{perrone2019learning} or hyperparameter importance \citep{hoos2014efficient}.

On the other hand, the ensuing problem is that when we need the early-stop techniques. Obviously, if there is no relationship between the performance with different budgets, these techniques will not work. Thus, how to judge whether a task is suitable for early stopping becomes an important prerequisite which this work is not involved in. We suggest to consider different criteria for different goals. For best arm identification, SH based algorithms have theoretical advantages. However, for the BO iteration, this goal is no longer appropriate which is replaced by the cumulative regret in this work. 


\newpage
\appendix
\section*{Proof of Theorem \ref{thm:two1} }\label{sec:proof}

First, we prove a lemma which is an extension of Theorem $1$ in \citet{Chan2019The},
\begin{align*}
\limsup_{r\rightarrow\infty}\mathbb E n_k^{(r)}/\log r&\leq\sum_{k:\mu_*>\mu_k}\phi_*/[(b(\mu_k)-b(\mu_*))\mu_k\\
&-(g(b(\mu_k))-g(b(\mu_*)))].
\end{align*}
For this purpose, the next main process is to develop a new Chernoff bound for the exponential family like Lemma $3$ in \citet{Chan2019The}. Now we claim that the following two inequalities hold for the exponential family. 

$$P(\bar Y^{(k)}_{1:j}\geq a) \leq \exp\{-jI_k(a)\}\qquad if\ a>\mu_k,$$
$$P(\bar Y^{(k)}_{1:j}\leq a) \leq \exp\{-jI_k(a)\}\qquad if\ a<\mu_k,$$

where the function $I_k(a)=\sup_{t\in \mathbb{R}} (ta-\ln\mathbb E \exp\{tY^{(k)}\})$ is the large deviation rate function.

The generic Chernoff bound for a random variable $\bar Y$ with $t>0$ is

$$P(\bar Y\geq a)=P(\exp\{t\bar Y\}\geq exp\{ta\})\leq \frac{\mathbb E \exp\{t\bar Y\}}{\exp\{ta\}}.$$

When $\bar Y=\bar Y^{(k)}_{1:j}$ is the mean of $j$ i.i.d. random variables $Y^{(k)}_1,\ldots,Y^{(k)}_j$, optimizing over $t$, we get

\begin{align*}
P(\bar Y^{(k)}_{1:j}\geq a)&\leq \inf_{t>0}\exp\{-ta\}\Pi_{i=1}^j\mathbb E \exp\{tY^{(k)}_i/j\}\\
&=\inf_{t>0}\exp\{-ta+\sum_{i=1}^j\ln\mathbb E \exp\{tY^{(k)}_i/j\}\}\\
&=\exp\{-j\sup_{t>0}(ta-\ln\mathbb E \exp\{tY^{(k)}_1\})\}\\
&=\exp\{-j\sup_{t>\theta_k}((t-\theta_k)a\\
&-\ln\mathbb E \exp\{(t-\theta_k)Y^{(k)}_1\})\}.
\end{align*}

Let $J_k(t)=(t-\theta_k)a-\ln\mathbb E \exp\{(t-\theta_k)Y^{(k)}_1\}$, we can simplify $J_k(t)$ by direct calculation that

$$J_k(t)=[(t-\theta_k)a-(g(t)-g(\theta_k))]/\phi_k.$$
Then,
$$J_k^{'}(t)=[a-g'(t)]/\phi_k,$$
and 
$$J_k^{''}(t)=-g''(t)/\phi_k<0.$$

Hence, when $a>\mu_k=g'(\theta_k)$, we have
$$\sup_{t>\theta_k}J_k(t)=J_k(b(a))=I_k(a),$$
which means the first inequality holds.

Similarly, when $a<\mu_k=g'(\theta_k)$, we have
$$\sup_{t<\theta_k}J_k(t)=J_k(b(a))=I_k(a).$$
The second inequality is obtained together with
\begin{align*}
P(\bar Y^{(k)}_{1:j}\leq a)&\leq \inf_{t<0}\exp\{-ta\}\Pi_{i=1}^j\mathbb E \exp\{tY^{(k)}_i/j\}\\
&=\exp\{-j\sup_{t<\theta_k}((t-\theta_k)a\\
&-\ln\mathbb E \exp\{(t-\theta_k)Y^{(k)}_1\})\}.
\end{align*}

Finally, let $\xi_k=1/I_*(\mu_k)$ in Lemma $1$ of \citet{Chan2019The}, we get an extension of Theorem $1$ in \citet{Chan2019The}. 
Consequently, the near optimality can be obtained as follows,
\begin{align*}
&\limsup_{N\rightarrow\infty} R(\hat\pi)/\log N\\
=&\limsup_{r\rightarrow\infty}\sum_{k:\mu_*<\mu_k}(\mu_k-\mu_*)\mathbb E n_k^{(r)}/\log n^{(r)}\\
\leq&\limsup_{r\rightarrow\infty}\sum_{k:\mu_*<\mu_k}(\mu_k-\mu_*)\mathbb E n_k^{(r)}/\log r\\
\leq&\sum_{k:\mu_*<\mu_k}(\mu_k-\mu_*)\phi_*/[(b(\mu_k)-b(\mu_*))\mu_k\\
&-(g(b(\mu_k))-g(b(\mu_*))).
\end{align*}

\QEDB

\vskip 0.2in
\bibliography{ref}

\end{document}